\documentclass[lettersize,journal]{IEEEtran}
\usepackage{amsmath,amsfonts}
\usepackage{algorithm}
\usepackage{array}
\usepackage[caption=false,font=normalsize,labelfont=sf,textfont=sf]{subfig}
\usepackage{textcomp}
\usepackage{stfloats}
\usepackage{url}
\usepackage{verbatim}
\usepackage{graphicx}
\usepackage{cite}
\usepackage{pgfplots}
\pgfplotsset{compat=1.18}
\usepackage{acronym}
\usepackage{xcolor}
\usepackage{multirow}
\usepackage{booktabs}
\usepackage{threeparttable}
\usepackage{makecell}
\usepackage{algpseudocode}
\usepackage{algorithm}

\algnewcommand\algorithmicparameters{\textbf{Parameters:}}
\algnewcommand\Parameters[1]{\item[\algorithmicparameters] #1}
 
\algnewcommand\algorithmicinput{\textbf{Input:}}
\algnewcommand\Input[1]{\item[\algorithmicinput] #1}
 
\algnewcommand\algorithmicoutput{\textbf{Output:}}
\algnewcommand\Output[1]{\item[\algorithmicoutput] #1}

\acrodef{LLM}{Large Language Model}
\acrodef{CNN}{Convolutional Neural Network}
\acrodef{RNN}{Recurrent Neural Network}

\begin{document}
\title{DeltaLLM: Delta-Key Sparse Attention for Edge LLMs}
\author{Jiawen Qi, Jiayu Liu, Chang Gao,~\IEEEmembership{Member,~IEEE}, Zhaochun Ren, Qinyu Chen,~\IEEEmembership{Senior Member,~IEEE}
\thanks{The authors acknowledge the Dutch Research Council (NWO) Veni Talent Program (21828) and the LIACS strategic PhD program for funding of this work. (\textit{Corresponding author: Qinyu Chen.)}}
\thanks{Jiawen Qi, Jiayu Liu, Zhaochun Ren, and Qinyu Chen are with the Leiden Institute of Advanced Computer Science (LIACS), Leiden University, 2300 RA Leiden, The Netherlands (e-mail: q.chen@liacs.leidenuniv.nl).}
\thanks{Chang Gao is with the Department of Microelectronics, the Delft University of Technology, 2628 CD Delft, The Netherlands.}
}





\maketitle

\begin{abstract}
On-device \ac{LLM} inference is becoming increasingly important for edge platforms, yet practical deployment remains challenging due to the high computational and memory demands of \acp{LLM} and the tight resource constraints of edge devices. 
Modern \ac{LLM} applications often involve long prompts. Although recent compact models, such as BitNet and LLaMA3.2, reduce the cost of linear layers through weight compression, they leave attention computation largely unoptimized, causing attention to overtake linear layers as the dominant bottleneck, especially in the prefilling stage. 
Existing attention pruning methods are primarily designed for highly parallel hardware (e.g., GPUs) and target long context windows, making them unsuitable for edge scenarios.
We present \textit{DeltaLLM}, a training-free \ac{LLM} framework based on algorithm-hardware-dataflow co-design to accelerate prefilling attention. 
\textit{DeltaLLM} introduces a causal-attention-aware K-only delta formulation. The formulation enables Delta-Key reuse across query heads under Grouped-Query Attention (GQA). A context-aware hybrid strategy further preserves exact attention near the diagonal while applying delta approximation to nonlocal regions. For score--value multiplication, we propose a vectorized structured sparse online softmax (VSSOS) that converts fine-grained spatial score sparsity into hardware-friendly column-level skipping. These algorithms are supported by a specialized vector-processor architecture and unified sparse/dense attention dataflow to fully exploit the sparsity brought out by our algorithm.
Evaluated on BitNet-b1.58-2B-4T and LLaMA3.2-1B-Instruct, \textit{DeltaLLM} achieves up to 60\% Delta-Key sparsity and 63\% structured attention-score sparsity, with negligible average accuracy loss.
On the KV260 FPGA board, it delivers 1.62x speedup over TeLLMe-v2 while reducing DSP and BRAM usage.
These results show that \textit{DeltaLLM} is well-suited for edge LLM deployment, offering integration with existing pipelines without fine-tuning.
\end{abstract}

\section{Introduction}
\label{sec:Intro}
\acp{LLM}, such as GPT~\cite{bubeck2023sparks} and LLaMA~\cite{touvron2023llama}, have demonstrated remarkable capabilities across a wide range of natural language processing tasks. They are typically deployed in cloud environments equipped with high-end GPUs (e.g., NVIDIA H100) or NPUs (e.g., Google TPU v7) due to their high computational and memory demands. Recently, there has been growing interest in bringing \acp{LLM} to edge devices to support offline applications, avoid communication latency, and preserve privacy. However, deploying \acp{LLM} on edge devices remains challenging, as they have far fewer computational resources, restricted memory bandwidth, and limited power budgets. 
These limitations necessitate effective \ac{LLM} cost-reduction techniques for on-device inference without compromising performance. 

\begin{figure}[t]
\centering
\includegraphics[width=1\columnwidth]{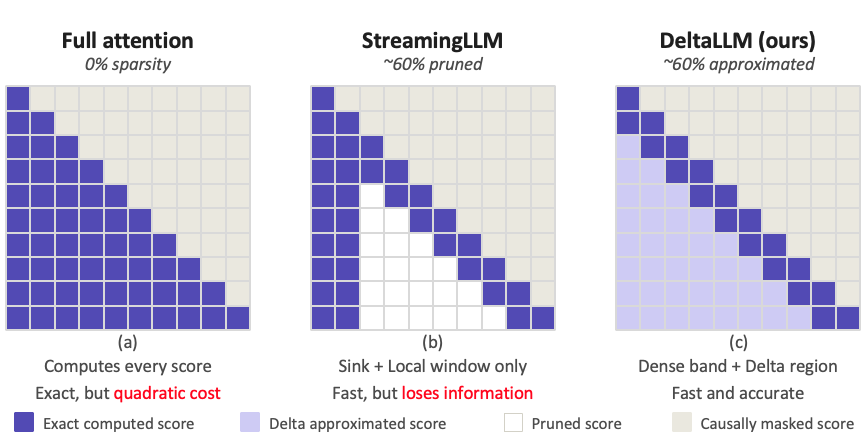}
\caption{Three different attention computation schemes. (a) Vanilla attention without exploiting any sparsity. (b) StreamingLLM~\cite{xiao2023efficient} computes attention following the \textbf{$\Lambda$-shape} pattern. (c) Our proposed method uses the temporal sparsity of key tokens to approximately compute most attention scores.}
\label{fig:1}
\end{figure}

Recent studies have exploited sparsity in \acp{LLM} to reduce the inference cost. 
For linear layers, trivial weights or activations are pruned~\cite{sun2023simple, liu2024training, Grasp2025DAC}.
However, as models shrink through weight compression and input length continues to grow, the computational bottleneck gradually shifts from the linear layers to the attention layer~\cite{fichtl2025}. 
For example, in the dense BitNet-b1.58-0.7B~\cite{wang2023bitnet}~\cite{bitnet_b1_58_large} prefilling stage, attention accounts for approximately 8.8\%, 27.7\%, 43.3\%, and 60.5\% of the combined attention-plus-linear-layer MACs at input lengths of 128, 512, 1{,}024, and 2{,}048, respectively.
To accelerate attention, dynamic sparse attention approaches such as MInference~\cite{jiang2024minference}, InfLLM~\cite{xiao2024infllm}, FlexPrefill~\cite{lai2025flexprefill}, OmniKV~\cite{hao2025omnikv}, and SpargeAttn~\cite{zhang2025spargeattn} identify significant tokens at runtime before calculating attention scores.  
The extra computational overhead required to find important key tokens can offset the benefit of sparsification in edge scenarios.
Static pruning methods are much simpler with almost no overhead. 
For example, StreamingLLM~\cite{xiao2023efficient} adopts a fixed token-selection \textbf{$\Lambda$-shape} pattern based on the observation that most attention scores are located within the attention sink (the first few columns) and the context window near the diagonal (Fig.~\ref{fig:1}(b)). 
However, this rigid pattern can degrade accuracy dramatically by discarding informative tokens outside predefined regions. 
These approaches are primarily motivated by very long contexts (i.e., 8k--128k), whereas edge applications often operate at moderate input lengths, where aggressive token removal is less tolerable.

Beyond algorithms, dedicated hardware architectures~\cite{VEDA2025DAC,zeng2024flightllm,EdgeLLM2025tcas1} have been proposed to speed up attention computation. 
They rely on abundant resources such as wide systolic arrays and high-bandwidth memory (HBM), incurring high area and power costs that are impractical for edge devices. 
Edge-oriented accelerators such as Hummingbird+~\cite{li2026hummingbird+}, together with recent large-model accelerators such as Panacea~\cite{kam2025panacea}, primarily accelerate low-precision or sparse GEMM operations in linear layers. TeLLMe-v2~\cite{qiao2026tellme} additionally incorporates a dedicated reverse prefilling attention (RPA) unit, providing attention-specific acceleration on the KV260 board. 
However, it does not exploit the attention sparsity.
Therefore, efficiently accelerating attention computation remains an open challenge for embedded devices.
We focus on the prefilling stage because it exposes a two-dimensional query--key attention map whose quadratic cost dominates edge inference for moderate-to-long prompts.
This separation is common in \ac{LLM} accelerator design, as prefilling and decoding exhibit distinct computational patterns and are often optimized with different algorithms and dataflows.
In contrast, decoding processes only one new query token at each step, reducing attention computation to a vector--matrix operation and shifting the bottleneck toward KV-cache access and weight loading.
Recent state-aware sparsity and quantization methods therefore compress the KV cache to reduce decoding-stage memory costs~\cite{yun2026asymmetric}. This direction is complementary to methods that accelerate the prefilling attention.



Our work, \textit{DeltaLLM}, takes the next step by exploiting sparsity in both key tokens and attention scores through algorithm--hardware co-design. 
There are two matrix multiplications in the attention layer: query--key multiplication and score--value multiplication.
For the query--key matrix multiplication, rather than pruning key tokens based on predicted or assumed positions of the salient attention scores, \textit{DeltaLLM} approximates most attention scores at a computational cost comparable to direct pruning, while preserving information from all token positions (Fig.~\ref{fig:1}(c)).
The approximated attention scores are computed through a delta-based mechanism~\cite{neil2017delta}.
It converts dense vectors into temporally sparse delta representations by computing differences between consecutive vectors and zeroing out small, uninformative changes.
Previous delta networks have been applied to recurrent and convolutional models~\cite{neil2017delta,gao2022spartus,habibian2021skip,Parger2022DeltaCNN,Parger2023MotionDeltaCNN,deltakws2025} and to a small encoder-style keyword Transformer~\cite{jelvcicova2022delta}. Directly applying a generic delta formulation to causal LLM prefilling, however, does not account for the lower-triangular causal mask, attention-sink and near-diagonal saliency, Grouped-Query Attention (GQA)-based key sharing, or block-wise online-softmax execution, leading to accuracy degradation. DeltaLLM instead introduces a causal-attention-aware K-only delta mapping: it preserves an exact first score column aligned with the attention sink and allows each generated Delta-Key head to be reused across all query heads in the same GQA group. A hybrid strategy further preserves exact attention near the diagonal while applying delta approximation only to nonlocal regions.

For score--value multiplication, although the score matrix is highly sparse, its fine-grained unstructured pattern is not aligned with hardware parallelism, hindering efficient computation.
DeltaLLM addresses this mismatch through vectorized structured sparse online softmax (VSSOS), which converts fine-grained score sparsity into column-level skipping during block-wise online softmax and score--value multiplication. These algorithms are co-designed with an edge-oriented architecture and a unified dataflow that supports full and delta attention blocks, streaming K/V data, and variable-length compressed Delta-Key vectors.
Our work makes the following contributions: 
\begin{itemize}
    \item We introduce a causal-attention-aware K-only delta mapping for query--key multiplication. By aligning the exact basis column with the attention sink, the mapping avoids the unfavorable interaction between query-side delta construction and causal masking, while enabling Delta-Key reuse across query heads under GQA.
    \item We develop a hybrid attention strategy that dynamically mixes full attention within a local context window and delta-based approximate attention outside the window to preserve accuracy.
    \item We develop VSSOS, which converts fine-grained attention-score sparsity into hardware-friendly column-level skipping during block-wise online softmax and score--value multiplication.
    \item We develop an edge-oriented hardware architecture with dedicated sparse key/score generators and a vector processing unit, together with a unified dataflow supporting full and delta attention blocks, streaming K/V data, and variable-length compressed Delta-Key vectors.
    \item Under acceptable accuracy loss on the BitNet model, our system achieves a 1.62$\times$ speedup of prefilling attention over the SOTA, with lower resource utilization and 2.9\,W power consumption on the KV260.
\end{itemize}

\section{Background}

\subsection{Attention Mechanism}
Attention enables \acp{LLM} to condition every token on all other tokens in the sequence to dynamically determine important information in the context. 
Given an input sequence $X\in\mathbb{R}^{n\times d}$, where $n$ is the number of tokens and $d$ is the token embedding length, it is linearly projected into queries $Q$, keys $K$, and values $V$ (each in $\mathbb{R}^{n\times d}$). In multi-head attention, they are reshaped into multiple query heads ($h_q$) along the hidden dimension, where each head has a shorter hidden dimension $d_{head}$. 
Each head computes the scaled dot-product attention:
\begin{align}
   S=\operatorname{Softmax}\left(\frac{QK^{\top}}{\sqrt{d_{head}}}\right), O=SV.
\end{align}
The attention score requires $\Theta(n^{2}d)$ operations, causing the cost to grow quadratically as the input length increases. On edge devices with limited compute and memory bandwidth, this quadratic term rapidly becomes the dominant bottleneck once sequences exceed a few hundred tokens.

\subsection{Delta Network Algorithm}
The Delta Network is a biologically inspired algorithm that converts dense matrix multiplication to temporally sparse matrix multiplication. Given two dense matrices $A$ and $B$, the dot product $R$ is computed by two steps: 

\textbf{Step 1: Delta Generation.} 
Each row (or column) of $A$ is treated as a temporal sequence $a(0)$, $a(1)$, \ldots, $a(t)$. 
The first row (or column) $a(0)$ is used as the basis vector, and the remaining rows (or columns) are represented by delta vectors $\Delta a(1), \Delta a(2), \ldots, \Delta a(t)$. 
We define $\Delta a(0)=a(0)$. For $t>0$, $\Delta a(t)$ is obtained by comparing the difference between the current vector $a(t)$ and the previous reference vector $\hat{a}(t-1)$ to a threshold $\theta$:

\begin{align}
\Delta a(t) &= 
\begin{cases}
a(t) - \hat{a}(t-1), & |a(t) - \hat{a}(t-1)| > \theta \\
0, & |a(t) - \hat{a}(t-1)| \leq \theta \\
\end{cases} \\
\hat{a}(t) &=
\begin{cases}
a(t), & |a(t) - \hat{a}(t-1)| > \theta \\
\hat{a}(t-1), & |a(t) - \hat{a}(t-1)| \leq \theta
\end{cases} 
\end{align}
The retained reference is initialized as $\hat{a}(0)=a(0)$, so the first vector is represented exactly.

\textbf{Step 2: Dense-delta matrix multiplication.} 
The output $R$ is computed recursively using the delta vectors.
Taking the initial output $R(0)$ as the basis vector, we calculate the output vectors $R(1), \ldots, R(t)$ recursively:
\begin{align}
    R(t) = 
\begin{cases}
    \Delta a(0)B, & t=0 \\
    \Delta a(t)B + R(t-1), &t >0 \\
\end{cases}
\end{align}
The output matrix is therefore the stack of $R(t)$.
Since $\Delta a(t)$ is often sparse, this formulation reduces the number of effective multiplications.

\section{DeltaLLM Algorithm}\label{sec:delta_contruct}
DeltaLLM algorithm introduces a causal-attention-aware K-only delta formulation, a context-aware hybrid strategy preserving exact attention near the diagonal while applying delta approximation to nonlocal regions and a vectorized structured sparse online softmax (VSSOS) algorithm converting fine-grained spatial score sparsity into hardware-friendly column-level skipping.
\subsection{Causal-Attention-Aware K-only Delta Mapping}

The target models apply rotary position embeddings (RoPE) to the projected queries and keys before attention. $Q$ and $K$ in this paper therefore denote the post-RoPE representations, and Delta-Key generation is applied to the post-RoPE keys. Accordingly, the reported Delta-Key sparsity is measured on the actual position-encoded keys used in $QK^{\top}$; RoPE itself is neither removed nor approximated.

In the prefilling stage, attention exhibits characteristic sparsity patterns where salient attention scores concentrate near the diagonal and in initial columns. 
Preserving these salient regions is crucial for maintaining model accuracy. 
Moreover, the causal attention mask restricts valid scores to the lower triangular region of the attention map. 
These structural properties must be considered when constructing Delta-Key matrices.
Naively applying the Delta Network to both the query and key matrices would introduce approximation errors uniformly across all attention scores, resulting in accuracy degradation.
To overcome this limitation, we develop a causal-attention-aware Delta-Key matrix construction strategy.

We observe that the choice of which matrix to transform (query or key) and which row is used as the reference (i.e., the order of computation) affects accuracy due to causal masking and the presence of the attention sink. 
We therefore empirically analyze three delta matrix construction strategies, as shown in Fig.~\ref{fig:3}:
\begin{figure}[t]
\centering
\includegraphics[width=1\columnwidth]{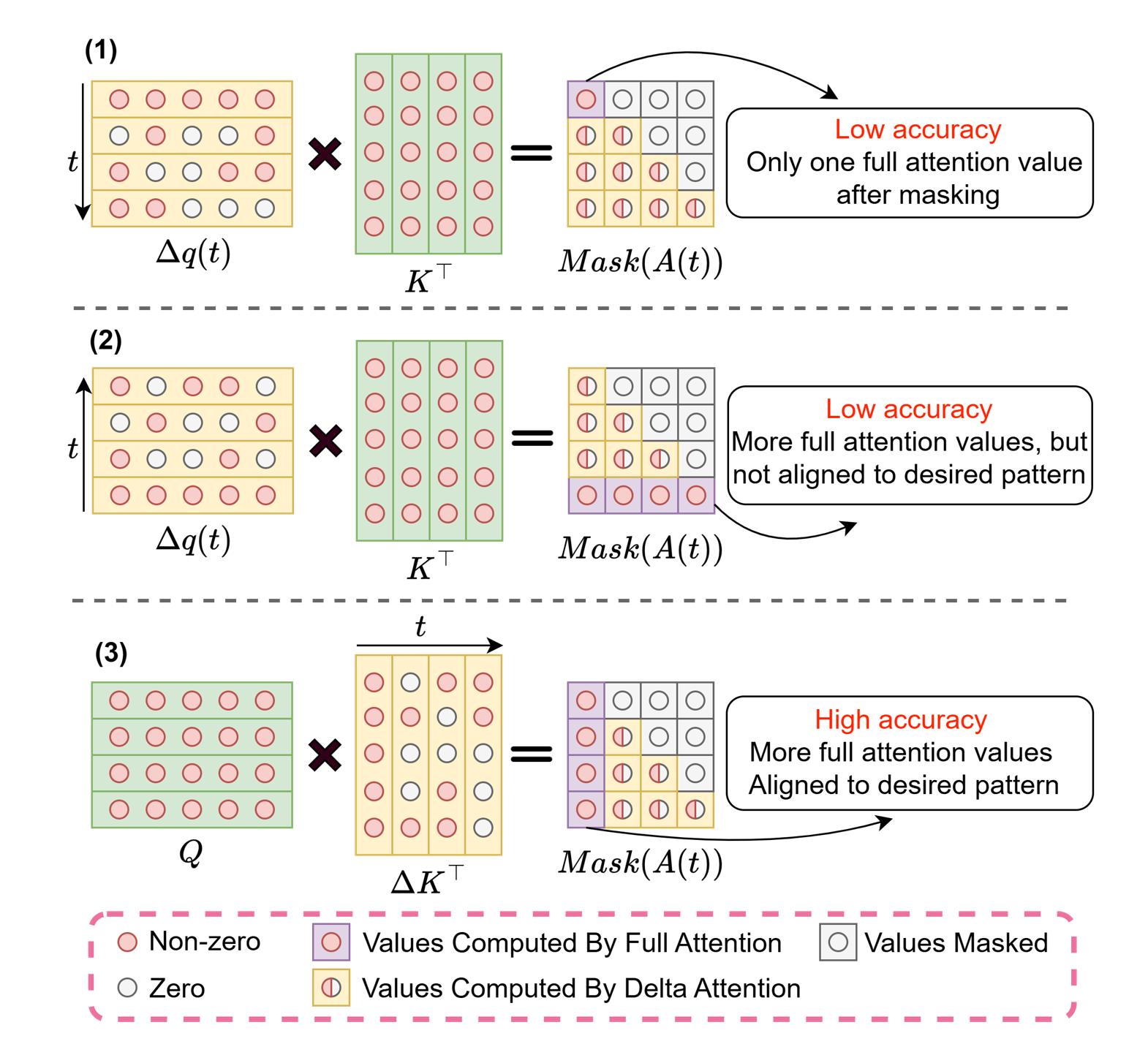}
\caption{Effect of delta construction direction on exact-score preservation under causal masking.}
\label{fig:3}
\end{figure}
\begin{itemize}
 \item[\textbf{(1)}] \textbf{Top-down $\Delta Q$ only:} Generate $\Delta q(t)$ starting from the top row of the query matrix while keeping keys unchanged. This produces a dense first row in the delta matrix, yielding exact computation for the first attention row. However, after masking, only the top-left attention score remains exactly accurate, resulting in poor performance.
 
 \item[\textbf{(2)}] \textbf{Bottom-up $\Delta Q$ only:} Generate $\Delta q(t)$ starting from the bottom row of the query matrix. This improves upon strategy (1) by preserving all accurate scores after masking, as the dense row aligns with the unmasked region.
 
 \item[\textbf{(3)}] \textbf{Top-down $\Delta K$ only (Ours):} Generate $\Delta k(t)$ starting from the top row of the key matrix while keeping queries unchanged. This preserves the entire first column of attention scores exactly, aligning with the attention sink pattern critical for model accuracy.
\end{itemize}

To quantitatively show the benefit of our causal-attention-aware top-down $\Delta K$ construction, we compare the accuracy of DeltaLLM on the HellaSwag dataset~\cite{zellers2019hellaswag} with the prior DKT-style construction~\cite{jelvcicova2022delta} that applies the delta rule to both $Q$ and $K$ matrices, and with the top-down and bottom-up query-only constructions under a matched computation budget (similar sparsity).

\begin{table}[t]
\centering
\caption{HellaSwag accuracy of delta-construction methods at comparable $QK^{\top}$ sparsity on BitNet-b1.58-2B-4T.}
\label{tab:delta_construction_comparison}
\footnotesize
\setlength{\tabcolsep}{2pt}
\renewcommand{\arraystretch}{1.10}

\begin{tabular*}{\columnwidth}{@{\extracolsep{\fill}}>{\raggedright\arraybackslash}p{0.43\columnwidth}cccc@{}}
\toprule
\textbf{Method}
& \textbf{$\theta_q$}
& \textbf{$\theta_k$}
& \textbf{$QK^{\top}$ Sp.}
& \textbf{HS Acc.} \\
\midrule
Dense                              & --  & --  & 0.00\%  & 68.45\% \\
DKT-style ($\Delta Q,\Delta K$)~\cite{jelvcicova2022delta}    & 0.2 & 0.7 & 53.42\% & 66.32\% \\
Top-down $\Delta Q$ only           & 0.6 & --  & 52.75\% & 65.38\% \\
Bottom-up $\Delta Q$ only          & 0.6 & --  & 52.98\% & 65.77\% \\
\hline
\textbf{Top-down $\Delta K$ only (Ours)} & --  & 1.0 & 53.31\% & \textbf{68.02\%} \\
\bottomrule
\end{tabular*}
\end{table}

As shown in Table~\ref{tab:delta_construction_comparison}, Top-down $\Delta K$ only construction achieves 68.02\% accuracy at 53.31\% sparsity, only 0.43 percentage points below the dense baseline. Under nearly identical sparsity, it outperforms the DKT-style, top-down query-only, and bottom-up query-only constructions by 1.70, 2.64, and 2.25 percentage points, respectively. These results demonstrate that the accuracy advantage does not arise merely from applying delta encoding; it depends on aligning the delta construction with the causal structure and salient regions of the attention map.

Guided by Fig.~\ref{fig:3} and Table~\ref{tab:delta_construction_comparison}, we adopt the top-down $\Delta K$ construction, using the key matrix as the delta source and the first key vector as the reference. This choice also reduces Delta-Key-generation overhead under GQA, which is used by recent \ac{LLM} architectures~\cite{grattafiori2024llama,wang2023bitnet,yang2025qwen3,GQA2023}. Because multiple query heads share the same key/value head, one Delta-Key matrix can be reused across the corresponding query heads, whereas DKT-style and query-only constructions additionally generate deltas for more numerous query heads.

\subsection{Bounded Delta-Key Error Propagation}
\label{sec:delta_error_bound}

DeltaLLM uses an error-feedback reference rather than independently thresholding consecutive differences. Let $\widehat{k}_t$ denote the retained reference after processing the post-RoPE key $k_t$, and define the reconstruction residual $e_t=k_t-\widehat{k}_t$. For each coordinate $m$, if $|[k_t-\widehat{k}_{t-1}]_m|>\theta_k$, the exact difference to the retained reference is emitted and the reference is refreshed to the current value, giving $[e_t]_m=0$. Otherwise, the previous reference is retained and the tested deviation is at most $\theta_k$. Therefore, at every token position,
{
\begin{align}
    |[e_t]_m| \leq \theta_k,
    \qquad
    \lVert e_t\rVert_\infty \leq \theta_k.
\label{eq:delta_residual_bound}
\end{align}
}
Discarded changes remain in the deviation from the last retained reference and are tested again at the next token; they are not summed as independently discarded adjacent differences. Hence, the residual is uniformly bounded in $t$ and cannot grow as $O(t\theta_k)$.

This bound also applies when Delta-Key generation follows RoPE. Let $q_i^R=R_iq_i$ and $k_j^R=R_jk_j$ be the encoded query and key, and let $\widehat{k}_j^R=k_j^R-e_j$. The approximate attention logit is
{
\begin{align}
\widehat{s}_{i,j}
&=\frac{(R_iq_i)^{\top}(R_jk_j-e_j)}{\sqrt{d_h}} \nonumber\\
&=\frac{q_i^{\top}R_{j-i}k_j}{\sqrt{d_h}}
-\frac{(R_iq_i)^{\top}e_j}{\sqrt{d_h}}.
\label{eq:rope_delta_error}
\end{align}
}
Thus, the original relative-position term $R_{j-i}$ remains intact. The logit perturbation can be bounded either by Cauchy--Schwarz or directly by Eq.~\ref{eq:delta_residual_bound}:
{
\begin{align}
\left|\widehat{s}_{i,j}-s_{i,j}\right|
&\leq \frac{\lVert q_i\rVert_2\lVert e_j\rVert_2}{\sqrt{d_h}},
\label{eq:rope_delta_bound}\\
\left|\widehat{s}_{i,j}-s_{i,j}\right|
&\leq \frac{\theta_k\lVert q_i^R\rVert_1}{\sqrt{d_h}}.
\label{eq:delta_logit_bound}
\end{align}
}

For a causal attention row, let $\epsilon_i=\lVert\widehat{s}_i-s_i\rVert_\infty$ over the valid positions $j\leq i$, and let $p_i=\operatorname{softmax}(s_i)$ and $\widehat{p}_i=\operatorname{softmax}(\widehat{s}_i)$. Softmax stability gives
{
\begin{align}
\lVert\widehat{p}_i-p_i\rVert_1
\leq 2\tanh(\epsilon_i/2)
\leq \epsilon_i.
\label{eq:delta_softmax_bound}
\end{align}
}
If $B_v=\max_{j\leq i}\lVert v_j\rVert_2$, the corresponding Delta-Key-induced attention-output perturbation satisfies
{
\begin{align}
\lVert\widehat{o}_i-o_i\rVert_2
\leq 2B_v\tanh(\epsilon_i/2).
\label{eq:delta_output_bound}
\end{align}
}
These bounds contain no term that grows with the input length. The residual of DeltaLLM is therefore bounded by \(\theta_k\) at every token position rather than being the sum of discarded adjacent differences, so the error caused by the quantization is not accumulated as the input length grows.




\subsection{Context-Aware Hybrid Block Attention}
\begin{figure}[t]
\centering
\includegraphics[width=1\columnwidth]{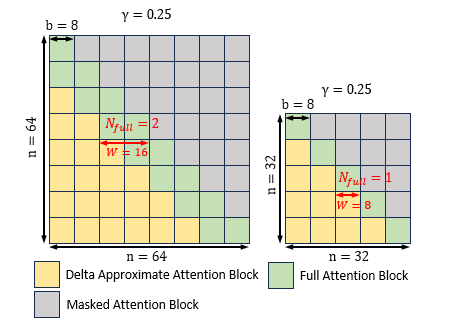}
\caption{Example of the context-aware hybrid attention computation pattern at input lengths $n=64$ and $n=32$, block size $b=8$, and scaling factor $\gamma = 0.25$.}
\label{fig:hybridAttn}
\end{figure}
\label{sec:ct_wd}

To improve accuracy while preserving sparsity, \textit{DeltaLLM} adopts a context-aware hybrid scheme: full attention, i.e., query and original key matrix multiplication, is applied within a local context window, while delta-based approximate attention, i.e., query and Delta-Key matrix multiplication, is performed outside the context window.
For a sequence of length $n$, we define a dynamic context window with length $W$:
\begin{align}
W = \min(\lfloor\gamma \cdot n\rfloor, W_{max})\label{eq:5}
\end{align}
where $\gamma \in (0,1)$ is a predefined scaling factor that dynamically controls the length of the context window according to the input length, and $W_{max}$ is an upper bound that prevents the context window from becoming too large.
The attention score is computed block-by-block rather than token-by-token, with each block containing $b$ tokens.
For a given query block, all key blocks must therefore be assigned either full attention or delta-based approximate attention.
Under the hybrid strategy, the number of key blocks computed with full attention for each query block is:
\begin{align}
    N_{full} = round(W/b),
\end{align}
while all remaining blocks for the same query block are evaluated using the delta-based approximate attention.

Fig.~\ref{fig:hybridAttn} shows an example of the full-attention and delta-approximate-attention pattern for different input lengths when the scaling factor is $\gamma=0.25$ and the block size is $b=8$.
The window size varies with the input length, resulting in fewer full-attention blocks for shorter inputs under the same scaling factor.

\begin{algorithm}[t]
\caption{Vectorized Structured Sparse Online Softmax}
\label{alg:ssos}
\begin{algorithmic}[1]
  \Parameters{$b_r, b_c$: query / key block sizes;\ $d_h$: head dim;\
           $T_s$: attention score threshold}
  \Input{$S \in \mathbb{R}^{b_r \times b_c}$: pre-softmax scores;\
           $V \in \mathbb{R}^{b_c \times d_h}$;\
           $m_{old}, l_{old} \in \mathbb{R}^{b_r}$;\
           $O_{old} \in \mathbb{R}^{b_r \times d_h}$}
  \Output{updated $O_{new},\ m_{new},\ l_{new}$}

  \Statex \textit{// Step 1: row-wise running max}
  \State $m_{new} \gets \max\!\big(m_{old},\ \operatorname{rowmax}(S)\big)$
         \Comment{vector length $b_r$}

  \Statex \textit{// Step 2: rescale factor for previously accumulated state}
  \State $\alpha \gets \exp(m_{old} - m_{new})$
         \Comment{vector length $b_r$}

  \Statex \textit{// Step 3: structured-sparse exponentiation, column-wise}
  \State $\beta \gets \mathbf{0} \in \mathbb{R}^{b_r \times b_c}$
  \For{$j \gets 0$ \textbf{to} $b_c - 1$}
    \State $\delta^j \gets S^{:,j} - m_{new}$
           \Comment{vector op, length $b_r$}
    \State $\mathit{mask}^j \gets (\delta^j \ge T_s)$
           \Comment{vector length $b_r$}
    \If{$\operatorname{any}(\mathit{mask}^j)$}
           \Comment{discard columns with all elements below the threshold}
      \State $\beta^{:,j} \gets \exp(\delta^j)$
    \EndIf
  \EndFor

  \Statex \textit{// Step 4: update denominator}
  \State $l_{new} \gets \alpha \odot l_{old} + \operatorname{rowsum}(\beta)$

  \Statex \textit{// Step 5: update partial output}
  \For{$j \gets 0$ \textbf{to} $d_h - 1$}
    \State $O_{new}^{:,j} \gets \alpha \odot O_{old}^{:,j}
            + \sum_{k=0}^{b_c-1} \beta^{:,k} \cdot V^{k,j}$
  \EndFor

  \State \Return $O_{new},\ m_{new},\ l_{new}$
\end{algorithmic}
\end{algorithm}

\begin{algorithm}[t]
\caption{DeltaLLM Algorithm}
\label{alg:overview}
\begin{algorithmic}[1]
  \Require $n$: total tokens;\ $b$: block size;\
           $\theta_k$: Delta-Key threshold;\ $T_s$: score threshold;\
           $\gamma$: window scale;\ $W_{\max}$: max window;\
           $d_h$: head dim;\ $h_{kv}$: \#KV heads;\
           $g$: GQA group size, $h_q = g \cdot h_{kv}$
  \Require $Q,\,K,\,V$ \Comment{$Q,K$ are post-RoPE}
  \Statex \hspace{\algorithmicindent}\textit{$Q$ has $h_q$ heads; $K,V$ each have $h_{kv}$ heads}
  \Ensure  attention output $O \in \mathbb{R}^{n \times h_q d_h}$

  \State $W \gets \min(\lfloor \gamma n \rfloor,\, W_{\max})$
  \State $N_{\text{full}} \gets \mathrm{round}(W / b),\quad N \gets \lceil n / b \rceil$
  \State $O \gets \mathbf{0}$

  \For{$h \gets 1$ \textbf{to} $h_{kv}$} \Comment{loop over KV heads}
    \State $\Delta K^{(h)} \gets \textsc{DeltaMatrixGen}(K^{(h)},\, \theta_k)$
        \Comment{Sec.~II-B; reused across $g$ query heads}
    \For{$g' \gets 1$ \textbf{to} $g$} \Comment{Q heads sharing $K^{(h)}, V^{(h)}$}
      \State $\hat{q} \gets (h - 1)\,g + g'$ \Comment{global Q-head index}
      \For{$i_q \gets 1$ \textbf{to} $N$} \Comment{Q block}
        \State $m \gets -\infty,\ \ l \gets 0,\ \ O_{\text{acc}} \gets \mathbf{0}$
            \Comment{online-softmax state}
        \State $a \gets \mathbf{0} \in \mathbb{R}^{b}$
            \Comment{carried last column of $A$ for delta accumulation}
        \For{$i_k \gets 1$ \textbf{to} $i_q$} \Comment{causal K-block loop}
          \If{$i_k \leq i_q - N_{\text{full}}$}
                \Comment{\textit{delta-attention region}}
            \State $\Delta A \gets \textsc{DenseSparseMM}\!\big(
                    Q_{i_q}^{(\hat{q})},\, \Delta K_{i_k}^{(h)\top}\big)$
            \State $A_{:,0} \gets a + \Delta A_{:,0}$
            \For{$\tau \gets 1$ \textbf{to} $b - 1$}
              \State $A_{:,\tau} \gets A_{:,\tau - 1} + \Delta A_{:,\tau}$
            \EndFor
            \State $a \gets A_{:,\,b-1}$ \Comment{carry to next delta block}
          \Else \Comment{\textit{full-attention region}}
            \State $A \gets \textsc{DenseMM}\!\big(
                    Q_{i_q}^{(\hat{q})},\, K_{i_k}^{(h)\top}\big)$
          \EndIf
          \State $S \gets A / \sqrt{d_h}$
          \If{$i_k = i_q$}
            \State apply causal mask to $S$
          \EndIf
          \State $(O_{\text{acc}},\, m,\, l) \gets
                  \textsc{VSSOS}(S,\, V_{i_k}^{(h)},\, m,\, l,\, O_{\text{acc}},\, T_s)$
        \EndFor
        \State $O\big[\text{block } i_q,\ \text{head } \hat{q}\big]
                \gets O_{\text{acc}} / l$
      \EndFor
    \EndFor
  \EndFor
  \State \Return $O$
\end{algorithmic}
\end{algorithm}
\subsection{Vectorized Structured Sparse Online Softmax}
Apart from query-key matrix multiplication, the other major computational cost in prefilling attention comes from score-value matrix multiplication.
For long prompts, storing the entire attention scores on chip is impractical. FlashAttention~\cite{dao2022flashattention} addresses this problem by decomposing the full attention matrix into smaller attention blocks and performing block-wise attention computation with online softmax, thereby reducing memory traffic.
However, FlashAttention assumes GPU-level compute density, large on-chip capacity, and high-bandwidth HBM, all unavailable on typical edge devices. 
More importantly, the FlashAttention algorithm does not exploit the effective sparsity in attention scores during score-value multiplication. In practice, the score matrix naturally contains more than 90\% negligible terms. Directly computing these terms is wasteful on resource-constrained edge devices.

To exploit this opportunity, we follow the block-wise computation formulation from FlashAttention but design a new vectorized structured sparse online softmax (VSSOS) algorithm. Within each block, VSSOS uses a magnitude-based threshold to identify columns whose exponential terms are all negligible, converting fine-grained score sparsity into structured column-level sparsity.

As shown in Algorithm~\ref{alg:ssos}, the kernel first updates the row-wise running maximum $m_{new} = \max(m_{old},\, \operatorname{rowmax}(S))$ and derives the rescale
factor $\alpha = \exp(m_{old} - m_{new})$, which compensates the previously
accumulated $l_{old}$ and $O_{old}$ for the shift in the maximum and thereby
preserves numerical stability across blocks.

Structured sparsity is introduced in the exponentiation stage. For each key
column $j$ of the block, we compute the centered scores
$\delta^{j} = S^{:,j} - m_{new}$ together with a predicate mask
$\mathit{mask}^{j} = (\delta^{j} \geq T_s)$, where $T_s$ is a small negative
log-domain threshold. The predicate is used only to determine whether the
column contains at least one non-negligible entry. When an entire column lies
below the threshold---i.e., $\operatorname{any}(\mathit{mask}^{j})$ evaluates
to \textit{false}---the column's exponential and the corresponding value
accumulation are skipped. Otherwise, we compute the complete column as
$\beta^{:,j}=\exp(\delta^{j})$; VSSOS does not perform lane-wise pruning
within a retained column.

The denominator is then updated as
$l_{new} = \alpha \odot l_{old} + \operatorname{rowsum}(\beta)$, and the
partial output as
$O_{new} = \mathrm{diag}(\alpha)\,O_{old} + \beta V$. We implement the latter
as a rank-$b_c$ outer-product accumulation along the head dimension $d_h$,
which maps naturally onto the vector processor: the $d_h$ axis is vectorized
to maximize lane utilization, while the $b_c$ loop carries the per-column
pruning, so every column skipped in the exponentiation stage also
eliminates its corresponding fused-multiply--add in the output update. All
conditional logic is expressed as predicated vector operations rather than
scalar branches, allowing the kernel to execute entirely within the vector
pipeline.

This method enables parallel computation with structured column skipping at a small block size.
The value of $T_s$ is determined by offline calibration on several different datasets. 
In our experiments, $T_s=-10$ provides over 60\% structured score sparsity with no observable accuracy degradation.
Compared with directly supporting arbitrary unstructured sparsity, the proposed column-level skipping requires no additional hardware resources and is well aligned with the proposed vector processor described in Section~\ref{sec:hardware}.

\subsection{Overall Workflow}\label{sec:alg_workflow}

Combining the strategies mentioned above, Algorithm~\ref{alg:overview} summarizes the complete workflow.
\textit{DeltaLLM} exploits the sparsity in both matrix multiplications in the attention layer.
The two optimizations are complementary: Delta-Key reduces the query--key path, after which score--value computation becomes a larger fraction of the remaining latency, and VSSOS targets this shifted bottleneck.
For the query-key multiplication, inputs are divided into $N$ blocks, which are either computed by delta attention using $\textsc{DenseSparseMM}$ or by full attention using $\textsc{DenseMM}$.
After that, the softmax operation and score-value matrix multiplication are computed by our VSSOS algorithm.
The final attention output is normalized by the denominator.

\begin{figure*}[t]
\centering
\includegraphics[width=\linewidth]
{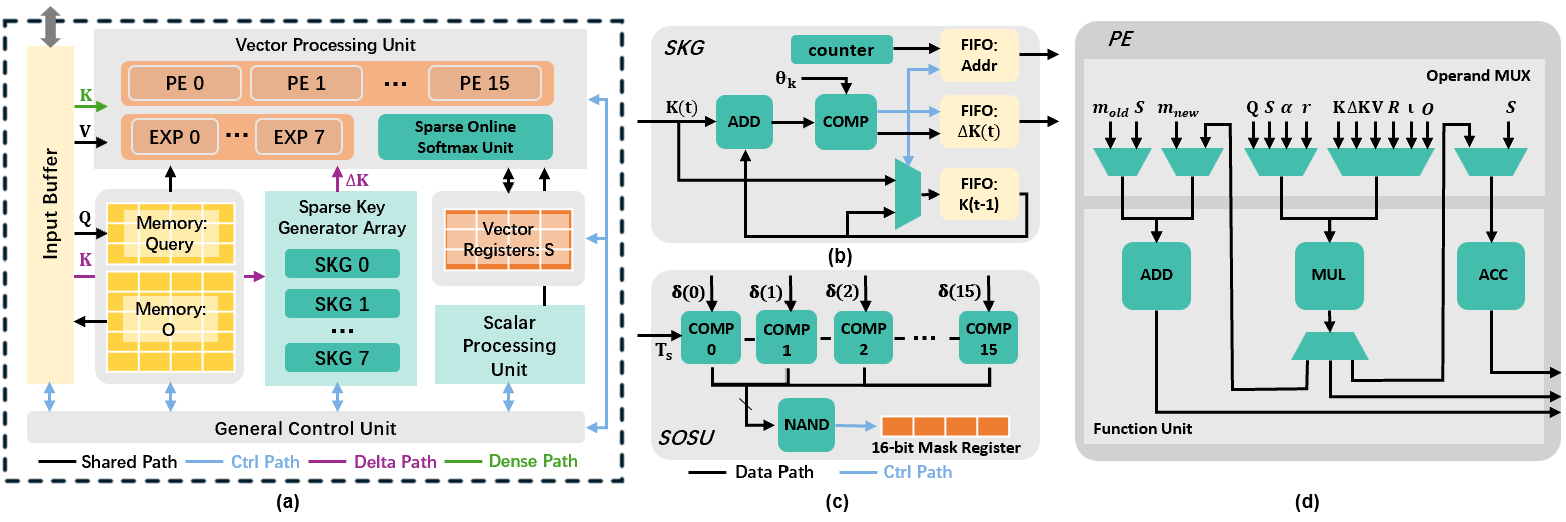}
\caption{(a) \textit{DeltaLLM} hardware architecture overview. $N_{pe}=16$. Microarchitecture of (b) SKG, (c) SOSU and (d) PE.}
\label{fig:hd_arch}
\end{figure*}

\section{DeltaLLM Hardware Architecture}
\label{sec:hardware}
\subsection{Architecture Overview}

Fig.~\ref{fig:hd_arch}(a) shows the overall hardware architecture of \textit{DeltaLLM} for accelerating prefilling attention. 
The design consists of several key components. 
First, a Sparse Key Generator (SKG) array applies the delta rule to generate fine-grained sparse key tokens and compress the sparse vectors into shorter dense key vectors.
Second, a Vector Processing Unit (VPU) is responsible for most of the computational load and performs vector operations, including multiplication, accumulation, addition, and exponentiation. 
Third, a Sparse Online Softmax Unit (SOSU) determines score columns whose entries are all zeros and tracks auxiliary state vectors for the online softmax operation. 
Additionally, the memory units contain on-chip memory for query tokens and attention outputs, as well as vector registers for attention scores and intermediate values.
A scalar processing unit is also used to calculate the reciprocal of the normalizers.
The general control unit controls the entire dataflow of the \textit{DeltaLLM} algorithm.
During prefilling, the \textit{DeltaLLM} architecture operates in two execution modes: 1) dense mode, which computes full attention for blocks within the context window, and 2) sparse mode, which applies delta-based approximate attention for the remaining blocks.
By combining the two modes, the architecture efficiently exploits temporal sparsity while maintaining accuracy.

\subsection{SKG Array}
The SKG array contains several SKG units. Each SKG unit produces sparse key tokens and records the indices of non-zero elements according to the delta rule.
As shown in Fig.~\ref{fig:hd_arch}(b), when a key element $K(t)$ is fetched from external memory, the corresponding retained reference $\widehat{K}(t-1)$ is read from the reference-key FIFO and subtracted from $K(t)$ to compute $\Delta K(t)$ through a 16-bit floating-point adder (ADD).
The resulting delta value is compared with a predefined threshold in the comparator (COMP) to generate a 1-bit control signal.
If the magnitude of the delta value is larger than the threshold ($\theta_k$), it is pushed into the $\Delta K(t)$ FIFO, while the remaining elements are discarded.
Meanwhile, the Addr FIFO records the position of the Delta-Key value in the original key token, and the reference-key FIFO updates the corresponding retained-reference element to form $\widehat{K}(t)$ based on the same control signal.
As a result, the long dense key vector is compressed into a shorter dense format. 
The number of SKGs is determined by the external memory bandwidth of the edge device. 
In our example, the bandwidth is 128 bits/cycle and the width of each key element is 16 bits, so the number of SKGs is $K = 128/16 = 8$. 
The FIFO depth $D$ is proportional to the hidden dimension of each attention head and is given by $D = d_{head} / K$.

\begin{figure*}[t]
\centering
\includegraphics[width=\linewidth]
{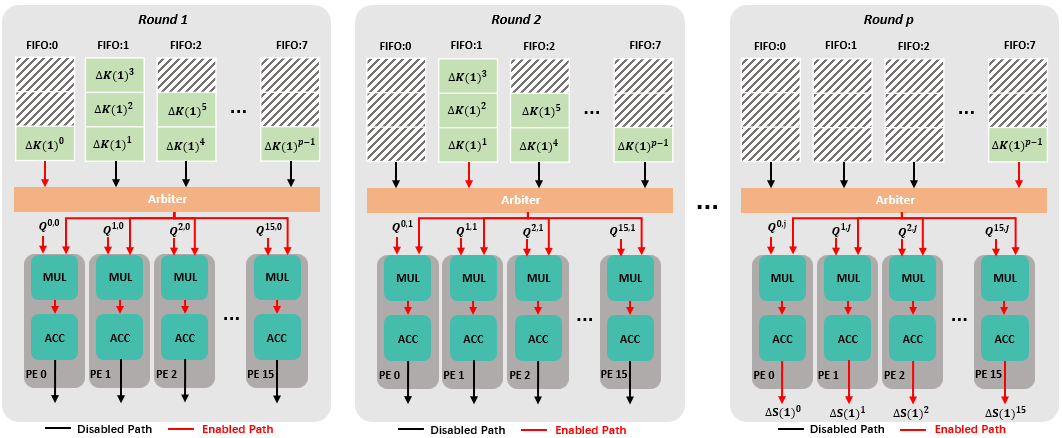}
\caption{Example dataflow of the query--Delta-Key dot product in \textit{DeltaLLM}.}
\label{fig:mm_flow}
\end{figure*}

After one entire token in the attention head is processed by the SKG array, the VPU will start to fetch the values from the FIFOs through an arbiter. 
As shown in Fig.~\ref{fig:mm_flow}, there is an arbiter used to control the dataflow between the SKG FIFOs, Query memory and the VPU. The pipeline of the arbiter is : (1) SKG FIFO selection; (2) Fetch the index from the selected ADDR FIFO; (3) Generate the corresponding address of the query column; (4) Read the query value and the delta key value; (5) Synchronization (6) Send the synchronized data to the VPU.

\subsection{SOSU}
The SOSU contains $N_{pe}$ comparators, a NAND gate, and a mask register.
The first and third steps from Algorithm~\ref{alg:ssos} are executed on the SOSU.
Fig.~\ref{fig:hd_arch}(c) shows the datapath in SOSU for the third step, where each centered score is compared with the predetermined threshold $T_s$ to generate $N_{pe}$ one-bit signals. If all comparator signals are '0's, the NAND gate generates a '1' and marks the corresponding column in the mask register. The mask register guides the VPU to skip attention-score columns whose entries are all below the threshold during sparse online softmax.

\subsection{VPU}
\label{sec:vpu}
Apart from the SOSU, the VPU consists of $N_{pe}$ PEs and $N_{pe}/2$ exponential processing elements to support vectorized computing. 
As illustrated in Fig.~\ref{fig:hd_arch}(d), each PE has an operand MUX unit and a function unit. 
Each function unit contains an FP16 multiplier (MUL), an FP16 adder (ADD), and an FP16 accumulator (ACC).
The operand multiplexer dynamically selects the appropriate inputs from the memory units and routes them to the functional units based on the current execution stage.
This enables the VPU to flexibly support all vector-level operations required by the attention pipeline, including \textsc{DenseSparseMM}, \textsc{DenseMM}, and the vector operations in \textsc{VSSOS}.

Traditional dot-product implementations, which rely on parallel multipliers followed by an adder tree, cannot exploit the unstructured sparsity produced by SKG.
Unused multipliers remain idle, resulting in low utilization and minimal speedup. 
Therefore, we redesign the dataflow of matrix multiplication.
Instead of reading a row vector from the first matrix and a column vector from the second matrix and performing element-wise multiplication followed by a reduction, we read a \textbf{column} vector from the first matrix and an \textbf{element} (scalar) from the column vector in the other matrix.
The scalar is broadcast and multiplied with the other vector to generate a partial result vector to be accumulated.
The next column vector from the first matrix is multiplied with the next scalar in the column vector until the end of the column.
For each multiplication, the previous partial result vector is accumulated with the new result.
Fig.~\ref{fig:mm_flow} illustrates an example dataflow of the query matrix multiplied with the Delta-Key vector ($\Delta K(1)$) generated by the second key token.
Since the number of non-zero elements varies dynamically, the compressed delta vectors stored in the SKG FIFOs have non-uniform lengths.

The other operand of the MUL is a query element from the query memory ($Q^{i,j}$, where $i$ is the row index and $j$ is the column index). 
The choice of $j$ for each issued element is guided by the indices from the Addr FIFO.
The output of MUL is sent to ACC to accumulate the partial sum.
After all $p_t$ nonzeros are processed, the correct output ($\Delta S(1)$) can be read from the ACC to the score vector register.
Through this dataflow, our VPU eliminates the adder tree and enables efficient processing of the sparse key tokens.

To further accelerate the computation and increase hardware utilization, we divide the vector operations involving multiplication into three groups: (1) multiplication and addition, (2) multiplication and accumulation, and (3) multiplication only.
The output of the multiplier is routed to the adder and accumulator so that they can be pipelined.
In addition, the number of exponential units is half the number of PEs to reduce resource utilization, since exponential operations account for only a small portion (about 5\%) of the total operations.

\subsection{Memory Units}
The memory units are partitioned into four parts: query memory (Q), output memory (O), attention score vector registers (S), and state vector registers (M).
The query memory stores the query tokens in an attention head.
There is no dedicated on-chip memory for complete key or value tensors: dense keys and values are streamed from external memory, while only transient Delta-Key and reference-key elements reside in the SKG FIFOs.
The output memory stores the final output for all attention heads and the partial attention output ($O_{old}$ in Algorithm~\ref{alg:ssos}).
The attention score vector registers save several different contents: (1) delta attention score ($\Delta S(t)$) (2) recovered delta attention score ($A(t)$) (3) dense attention score ($A$) (4) normalized attention score ($S$) (5) $\delta$ (6) exponential scores ($\beta$).
There is no data dependency among these contents, so they can be stored in a set of vector registers.
The state vector registers contain six auxiliary vectors used in the online softmax and normalization: (1) $m_{old}$ (2) $m_{new}$ (3) $l_{new/old}$ (4) $\alpha$ (5) intermediate vector ($\sum_{k=0}^{b-1} \beta^{k} * V^{k,j}$) (6) denominator: $1/l_{new}$ or $1/\sqrt(d_h)$.

Our hardware design is scalable with the size of the model. 
Given a model with the number of query heads $h_q$, number of key-value heads $h_{kv}$, number of queries in a key-value head $g=h_q/h_{kv}$, head hidden dimension $d_{head}$, element bit width $w$, and block input token size $b$, the number of SKGs can be scaled with the external bandwidth (BW) to maximize parallelization: $K = BW/w$.
The number of PEs in the VPU can also be scaled up with the block size $N_{pe} = g * b$ to support tasks with longer inputs.
The on-chip memory is partitioned into four regions (Q, O, S, M), with details described in Table~\ref{tab:sram}.

{\small
\begin{table}[t]
\renewcommand{\arraystretch}{0.7}  
\centering
\caption{On-chip Memory Partitioning}
\label{tab:sram}
\begin{tabular}{c c c}
\toprule
\textbf{Mem} & \textbf{Content} & \textbf{Size (bits)} \\
\midrule
Q & Queries for all query groups & $g*b*d_{head}*w$ \\
O & Attention Output & $g*b*d*w$ \\
S & Delta or full attention scores & $g*b^2*w$ \\
M & State vector registers & $6*g*b*w$ \\
\bottomrule
\end{tabular}
\end{table}
}

\subsection{Decode-Stage Support and System Boundary}
The proposed architecture primarily targets prefilling attention. During prefill, the dense $K$ and $V$ tensors used to construct the conventional KV cache remain in external memory, whereas the compressed $\Delta K$ vectors generated by the SKG are transient intermediates stored only in the SKG FIFOs. DeltaLLM therefore neither replaces nor modifies the dense KV-cache format, and decoding can begin without cache-format conversion.

The existing VPU also provides a functionally correct dense decode fallback. At each decode step, the newly generated query token is mapped to a partially occupied query block. The SKG and score-pruning paths are bypassed, while cached $K$ and $V$ blocks are streamed from external memory through the VPU to compute exact query--key products, online softmax, and score--value accumulation. Prefill and decode can consequently time-share the multipliers, accumulators, exponential units, state registers, and scalar normalization unit.

\section{Experimental Results}
\subsection{Experiment Setup}
\label{sec:setup}
We evaluate our \textit{DeltaLLM} algorithm on two representative edge-friendly \acp{LLM}: the LLaMA3.2-1B-Instruct model~\cite{grattafiori2024llama} and BitNet-b1.58-2B-4T~\cite{wang2023bitnet}.
BitNet is a ternary-weight model, designed for low-power, resource-constrained hardware, making it an ideal platform for assessing sparsity-driven optimizations.
Beyond simple perplexity tests, we evaluate zero-shot accuracy on a range of language understanding and reasoning tasks: ARC-Easy~\cite{allenai:arc}, ARC-Challenge~\cite{allenai:arc}, BoolQ~\cite{clark2019boolq}, 
HellaSwag~\cite{zellers2019hellaswag}, OpenbookQA~\cite{OpenBookQA2018}, PIQA~\cite{bisk2020piqa}, and Winogrande~\cite{sakaguchi2021winogrande}.
We use an LLM benchmark suite \textbf{lm-evaluation-harness}\footnote{lm-evaluation-harness, \url{https://zenodo.org/records/12608602}} with default settings on a single NVIDIA A100 GPU (40GB GPU memory) to evaluate the inference accuracy.
Importantly, no fine-tuning is performed.
\textit{DeltaLLM} operates as a purely inference-time optimization.

For the resource-matched performance comparison between DeltaLLM and the traditional GPU-based algorithms, we develop a trace-driven cycle-level common vector engine (CVE) simulator in Python. StreamingLLM, MInference, and the generic DeltaLLM variants use the same FP16 precision, vector width, number of exponential units, external-memory bandwidth, attention-block size, and on-chip buffering assumptions. The complete DeltaLLM configuration uses the same compute and memory resource budget but enables the proposed specialized dataflow and SKG/VPU/SOSU architecture. The cycle costs of all modeled operations are calibrated against our DeltaLLM RTL implementation. We feed the exact sparse masks or nonzero traces generated by each method to the simulator, including runtime pattern-generation costs where applicable. The configurations are selected to target approximately 65\% sparsity, and all reported speedups are normalized to dense full attention executed on the CVE.

For hardware evaluation, we implement RTL code for the \textit{DeltaLLM} hardware architecture on the KV260 FPGA, which is representative of LPDDR-based embedded SoCs and low-power FPGA platforms and has 1--2 orders of magnitude lower bandwidth than high-end designs such as FlightLLM~\cite{zeng2024flightllm}, EdgeLLM~\cite{EdgeLLM2025tcas1}, and XShift~\cite{xshift2025dac}.
We evaluate our system in terms of speedup, FPGA resource utilization, and power at 250\,MHz, and compare our results with the SOTA prefilling-attention implementation. 
The evaluation is performed on the BitNet-b1.58-0.7B~\cite{wang2023bitnet}~\cite{bitnet_b1_58_large} model which is also used in TeLLMe-v2~\cite{qiao2026tellme} for fair comparison.

\subsection{Accuracy Analysis}
We first analyze the impact of two key hyperparameters in \textit{DeltaLLM} for the query-key matrix multiplication: the Delta-Key threshold $\theta_k$, which controls the sparsity of the constructed Delta-Key matrix, and the context-window scaling factor $\gamma$, which determines the size of the full-attention region during prefilling.

{\small
\begin{table*}[ht]
\renewcommand{\arraystretch}{0.5}  
  \centering
    \caption{\textit{DeltaLLM} accuracy (\%) and sparsity (\%). Various $\theta_k$, $\gamma = 0.05$. ARCe, ARCc, BQ, HS, OQ, PQ, and WG denote ARC-Easy, ARC-Challenge, BoolQ, HellaSwag, OpenBookQA, PIQA, and WinoGrande, respectively.}
  \begin{tabular}{@{} l|c c|c c|c c|c c|c c|c c|c c @{}} 
    \toprule
    \multicolumn{1}{c}{}
      & \multicolumn{2}{c}{ARCe} & \multicolumn{2}{c}{ARCc}& \multicolumn{2}{c}{BQ} & \multicolumn{2}{c}{HS} & \multicolumn{2}{c}{OQ} & \multicolumn{2}{c}{PQ} & \multicolumn{2}{c}{WG} \\
    \cmidrule(lr){2-3}
    \cmidrule(lr){4-5}
    \cmidrule(lr){6-7}
    \cmidrule(lr){8-9}
    \cmidrule(lr){10-11}
    \cmidrule(lr){12-13}
    \cmidrule(lr){14-15}
    & {$Acc_n$} & {$S_p$} & {$Acc_n$} & {$S_p$} & {$Acc_n$} & {$S_p$} & {$Acc_n$} & {$S_p$} & {$Acc_n$} & {$S_p$} & {$Acc_n$} & {$S_p$} & {$Acc_n$} & {$S_p$} \\
    \midrule
    LLaMA (baseline)        & \textbf{63.25} & 0.00 & \textbf{37.88} & 0.00 & \textbf{69.11} & 0.00 &  \textbf{60.80} & 0.00 & \textbf{34.40} & 0.00 & \textbf{74.04} & 0.00 &  \textbf{59.90} & 0.00\\
    DeltaLLM ($\theta_k=0.6$)    & 62.84 & 37.78 & 37.46 & 36.85 & 68.84 & 42.83 & 60.33 & 40.18 & 34.00 & 31.86 & 72.75 & 36.75 & 57.70 & 39.87\\
    DeltaLLM ($\theta_k=0.8$)    & 61.62 & 45.42 & 36.10 & 46.45 & 68.32 & 52.62 & 59.68 & 49.56 & 33.80 & 39.92 & 73.51 & 45.65 & 58.33 & 48.85 \\
    DeltaLLM ($\theta_k=1.0$)      & 61.03 & 52.88 & 36.10 & 54.70 & 67.34 & 60.68 & 58.49 & 57.45 & 35.20 & 46.51 & 72.64 & 52.55 & 55.96 & 56.04\\
    \midrule
    BitNet (baseline)  & \textbf{75.00} & 0.00 & 49.65 & 0.00 &  \textbf{80.48}& 0.00 & \textbf{68.45} & 0.00 & 41.80 & 0.00 & 76.98 & 0.00 & \textbf{72.45} & 0.00 \\
    DeltaLLM ($\theta_k=1.0$)   & 74.54 & 46.83 & 49.23 & 48.97 & 79.63 & 58.63 & 68.02 & 53.31 & \textbf{42.00} & 46.34 & 76.83 & 47.41 & 71.59 & 56.33 \\
    DeltaLLM ($\theta_k=1.2$)   & 74.11 & 53.01 & \textbf{50.17} & 56.06 & 79.64 & 65.40 & 67.65 & 60.10 & 40.80 & 52.04 & \bfseries 77.10 & 54.49 & 71.43 & 62.98 \\
    DeltaLLM ($\theta_k=1.4$)   & 74.54 & 57.82 & 49.06 & 61.49 & 79.33 & 70.59 & 67.25 & 65.38 & 40.60 & 56.28 & 76.44 & 60.31 & 70.56 & 68.00 \\
    \bottomrule
  \end{tabular}
  \label{tab:1}
\end{table*}
}

{\small
\begin{table*}[ht]
\renewcommand{\arraystretch}{0.5}  
  \centering
    \caption{\textit{DeltaLLM} accuracy (\%) and sparsity (\%). Various $\gamma$, $\theta_k = 0.6$ for LLaMA and $\theta_k = 1.0$ for BitNet.}
  \begin{tabular}{@{} l|c c|c c|c c|c c|c c|c c|c c @{}} 
    \toprule
    \multicolumn{1}{c}{}
      & \multicolumn{2}{c}{ARCe} & \multicolumn{2}{c}{ARCc}& \multicolumn{2}{c}{BQ} & \multicolumn{2}{c}{HS} & \multicolumn{2}{c}{OQ} & \multicolumn{2}{c}{PQ} & \multicolumn{2}{c}{WG} \\
    \cmidrule(lr){2-3}
    \cmidrule(lr){4-5}
    \cmidrule(lr){6-7}
    \cmidrule(lr){8-9}
    \cmidrule(lr){10-11}
    \cmidrule(lr){12-13}
    \cmidrule(lr){14-15}
    & {$Acc_n$} & {$S_p$} & {$Acc_n$} & {$S_p$} & {$Acc_n$} & {$S_p$} & {$Acc_n$} & {$S_p$} & {$Acc_n$} & {$S_p$} & {$Acc_n$} & {$S_p$} & {$Acc_n$} & {$S_p$} \\
    \midrule
    LLaMA (baseline)        & \bfseries 63.25 & 0.00 & 37.88 & 0.00 & \bfseries 69.11 & 0.00 & \bfseries 60.80 & 0.00 & 34.40 & 0.00 & \bfseries 74.04 & 0.00 & \bfseries 59.90 & 0.00\\
    DeltaLLM ($\gamma=0.1$)     & 62.97 & 36.78 & \bfseries 38.57 & 33.00 & 68.87 & 40.32 & 60.51 & 37.47 & \bfseries 35.00 & 31.86 & 72.58 & 32.25 & 58.57 & 36.79\\
    DeltaLLM ($\gamma=0.05$)    & 62.84 & 37.78 & 37.46 & 36.85 & 68.84 & 42.83 & 60.33 & 40.18 & 34.00 & 31.86 & 72.75 & 36.75 & 57.70 & 39.87\\
    DeltaLLM ($\gamma=0.02$)    & 62.25 & 40.35 & 37.38 & 36.85 & 68.11 & 44.20 & 60.03 & 42.67 & 34.80 & 31.86 & 72.42 & 36.75 & 57.62 & 42.60 \\
    \midrule
    BitNet (baseline)  & \bfseries 75.00 & 0.00 & 49.65 & 0.00 & \bfseries 80.48 & 0.00 & \bfseries 68.45 & 0.00 & 41.80 & 0.00 & 76.98 & 0.00 & 72.45 & 0.00 \\
    DeltaLLM ($\gamma=0.1$)   & 74.74 & 46.82 & \bfseries 49.92 & 43.86 & 79.82 & 55.10 & 68.06 & 49.61 & \bfseries 42.40 & 46.33 & \bfseries 77.04 & 41.37 & 71.67 & 52.27 \\
    DeltaLLM ($\gamma=0.05$)  & 74.54 & 46.83 & 49.23 & 48.97 & 79.63 & 58.63 & 68.02 & 53.31 &  42.00 & 46.34 & 76.83 & 47.41 & 71.59 & 56.33 \\
    DeltaLLM ($\gamma=0.02$)   & 74.50 & 51.40 & 49.75 & 48.98 & 79.73 & 70.59 & 68.03 & 60.48 & 41.40 & 46.34 & 76.88 & 47.41 & \bfseries 72.54 & 60.20 \\
    \bottomrule
  \end{tabular}
  \label{tab:2}
\end{table*}
}

Table~\ref{tab:1} reports the effect of varying $\theta_k$ with $\gamma=0.05$ and $T_s=-10$.
The LLaMA3.2-1B-Instruct model exhibits accuracy degradation as $\theta_k$ increases, indicating higher sensitivity to aggressive sparsification.
In contrast, BitNet-b1.58-2B-4T is noticeably more robust: with $\theta_k=1.2$, it achieves 57.73\% average Delta-Key sparsity while its average accuracy decreases by only 0.56 percentage points, from 66.40\% to 65.84\%. 

Table~\ref{tab:2} provides a complementary view by varying $\gamma$ while fixing $\theta_k$ (0.6 for LLaMA and 1.0 for BitNet).
Individual task accuracies fluctuate slightly as $\gamma$ varies, while larger effective context windows generally improve aggregate accuracy by allocating more computation to full attention, at the cost of lower sparsity.
At $\gamma=0.02$, DeltaLLM reaches 60.2\% sparsity on WG while keeping its accuracy close to the dense baseline (72.54\% versus 72.45\%).

Each row in Table~\ref{tab:1} uses one model-level Delta-Key threshold shared across all seven tasks without per-task tuning; Table~\ref{tab:2} similarly fixes $\theta_k=1.0$ for BitNet-b1.58-2B and $\theta_k=0.6$ for LLaMA3.2-1B.

\begin{table}[t]
\centering
\caption{Effect of Delta-Key placement relative to RoPE on BoolQ ($\gamma=0.05$; $\theta_k=0.6/1.0$ for LLaMA/BitNet).}
\label{tab:rope_placement}
\begingroup
\small
\setlength{\tabcolsep}{4pt}
\renewcommand{\arraystretch}{1.08}
\begin{tabular*}{\columnwidth}{@{\extracolsep{\fill}}llcc@{}}
\toprule
\textbf{Model} & \textbf{Delta Placement} & \textbf{Acc. (\%)} & \textbf{Key Sp. (\%)} \\
\midrule
LLaMA3.2-1B & Before RoPE & 68.32 & 44.15 \\
             & After RoPE  & \textbf{68.84} & 42.83 \\
\midrule
BitNet-b1.58-2B & Before RoPE & 79.01 & 57.85 \\
                & After RoPE  & \textbf{79.63} & 58.63 \\
\bottomrule
\end{tabular*}
\endgroup
\end{table}

To isolate the interaction between Delta-Key generation and positional encoding, we additionally compare applying delta construction immediately before and after RoPE on BoolQ.
As shown in Table~\ref{tab:rope_placement}, post-RoPE Delta-Key generation improves accuracy by 0.52 and 0.62 percentage points on LLaMA and BitNet, respectively, while the sparsity difference between the two placements is at most 1.32 percentage points. The effectiveness of Delta-Key is therefore not sensitive to whether the approximation is introduced immediately before or after positional encoding. We retain the post-RoPE formulation because it operates directly on the key representation consumed by attention and naturally supports the recursive score recovery in Algorithm~\ref{alg:overview}.




Table~\ref{tab:deltallm_minference_streaming_avg} compares DeltaLLM with StreamingLLM~\cite{xiao2023efficient} and MInference~\cite{jiang2024minference} under the same seven zero-shot lm-evaluation-harness tasks on LLaMA-3.2-1B-Instruct and BitNet-b1.58-2B-4T.
For StreamingLLM, we use the sink-plus-sliding-window pattern with four attention-sink tokens and solve the window size per sample to match the target sparsity; RoPE positions are re-indexed following the StreamingLLM prefill formulation. 
For MInference, since the official high-level patch does not support either target model, we use the official raw sparse-attention kernels through a custom adapter; calibration selects the vertical-slash pattern for all layers.

Overall, DeltaLLM provides the best accuracy--sparsity trade-off. 
StreamingLLM suffers a large accuracy degradation even at moderate sparsity, dropping from the dense baseline by 6.21--12.34 percentage points on LLaMA and 9.44--18.95 percentage points on BitNet. MInference is substantially stronger than StreamingLLM, but DeltaLLM generally retains higher accuracy at comparable real sparsity. On BitNet, DeltaLLM achieves 65.84\% average accuracy at 57.73\% real sparsity, outperforming MInference at 59.92\% real sparsity by 1.43 percentage points and StreamingLLM at 54.72\% real sparsity by 18.57 percentage points. 
On LLaMA, DeltaLLM also improves over StreamingLLM by 7--10 percentage points across the matched sparsity range, while remaining competitive with MInference; at the highest sparsity setting, DeltaLLM and MInference obtain similar average accuracy, but DeltaLLM does so with a different temporal-spatial sparsity mechanism rather than relying on a fixed vertical-slash attention pattern. 
These results indicate that DeltaLLM preserves task accuracy more effectively than fixed sink-window sparsity and is especially advantageous on BitNet, where it consistently achieves smaller accuracy loss than MInference at similar sparsity levels.

\begin{table*}[!t]
\centering
\caption{Seven-task average accuracy and real sparsity, with $\Delta$Acc from the dense baseline in percentage points (\%).}
\label{tab:deltallm_minference_streaming_avg}
\scriptsize
\setlength{\tabcolsep}{3.5pt}
\renewcommand{\arraystretch}{1.08}
\begin{tabular*}{\textwidth}{@{\extracolsep{\fill}}llccc llccc@{}}
\toprule
\multicolumn{5}{c}{\textbf{LLaMA-3.2-1B-Instruct}} &
\multicolumn{5}{c}{\textbf{BitNet-b1.58-2B-4T}} \\
\cmidrule(lr){1-5}\cmidrule(lr){6-10}
\textbf{Method} & \textbf{Setting} & \textbf{Real Sp. (\%)} & \textbf{Acc. (\%)} & \textbf{$\Delta$Acc (\%)} &
\textbf{Method} & \textbf{Setting} & \textbf{Real Sp. (\%)} & \textbf{Acc. (\%)} & \textbf{$\Delta$Acc (\%)} \\
\midrule
StreamingLLM &
target $=0.30$ &
29.24 & 50.85 & $-6.21$ &
StreamingLLM &
target $=0.30$ &
29.24 & 56.78 & $-9.44$ \\

MInference &
target $=0.30$ &
29.98 & 56.00 & $-1.06$ &
MInference &
target $=0.30$ &
29.98 & 65.38 & $-0.84$ \\

DeltaLLM (Ours) &
$\theta_k{=}0.6,\gamma{=}0.1$ &
35.50 & 56.72 & $-0.33$ &
DeltaLLM (Ours) &
$\theta_k{=}1.0,\gamma{=}0.1$ &
47.91 & 66.24 & $-0.17$ \\

\midrule
StreamingLLM &
target $=0.45$ &
43.13 & 47.62 & $-9.43$ &
StreamingLLM &
target $=0.45$ &
43.13 & 51.19 & $-15.03$ \\

MInference &
target $=0.45$ &
45.16 & 55.66 & $-1.40$ &
MInference &
target $=0.45$ &
45.16 & 64.87 & $-1.35$ \\

DeltaLLM (Ours) &
$\theta_k{=}0.8,\gamma{=}0.05$ &
46.92 & 55.91 & $-1.15$ &
DeltaLLM (Ours)&
$\theta_k{=}1.0,\gamma{=}0.1$ &
47.91 & 66.24 & $-0.17$ \\

\midrule
StreamingLLM &
target $=0.60$ &
54.72 & 44.72 & $-12.34$ &
StreamingLLM &
target $=0.60$ &
54.72 & 47.27 & $-18.95$ \\

MInference &
target $=0.60$ &
59.92 & 55.31 & $-1.75$ &
MInference &
target $=0.60$ &
59.92 & 64.41 & $-1.81$ \\

DeltaLLM (Ours)&
$\theta_k{=}1.0,\gamma{=}0.05$ &
54.40 & 55.25 & $-1.80$ &
DeltaLLM (Ours)&
$\theta_k{=}1.2,\gamma{=}0.05$ &
57.73 & 65.84 & $-0.56$ \\
\bottomrule
\end{tabular*}
\vspace{2pt}
\begin{minipage}{0.98\textwidth}
\footnotesize
\end{minipage}
\end{table*}

\subsection{Compatibility with Weight Compression}

DeltaLLM is complementary to conventional weight-compression techniques at the operator level. Weight pruning and low-bit weight quantization primarily reduce the storage and computation of the Q/K/V/O projections and feed-forward layers, whereas DeltaLLM operates on runtime attention tensors and reduces the computation of $QK^{\top}$ and $SV$ through Delta-Key and structured score sparsity. The two types of optimization can therefore be applied jointly without modifying the DeltaLLM attention dataflow.

Our existing BitNet-b1.58-2B-4T results already represent an aggressive ternary-weight setting. To broaden the evaluation, we test LLaMA3.2-1B-Instruct using the original BF16 model, INT8 round-to-nearest (RTN) quantization, 4-bit AWQ~\cite{lin2024awq}, and 50\% Wanda pruning~\cite{sun2023simple}. 
We choose $\theta_k = 0.6$ for all weight compression experiments for fair comparison.

\begin{table*}[t]
\centering
\caption{DeltaLLM accuracy change from different weight compression techniques and corresponding Delta-Key sparsity on LLaMA3.2-1B-Instruct model.}
\label{tab:weight_compression}
\begingroup
\scriptsize
\setlength{\tabcolsep}{3pt}
\renewcommand{\arraystretch}{1.08}
\begin{tabular*}{\textwidth}{@{\extracolsep{\fill}}lcccccccc@{}}
\toprule
\multirow{2}{*}{\textbf{Task}} &
\multicolumn{2}{c}{\textbf{BF16}} &
\multicolumn{2}{c}{\textbf{INT8 RTN}} &
\multicolumn{2}{c}{\textbf{AWQ W4}} &
\multicolumn{2}{c}{\textbf{Wanda 50\%}} \\
\cmidrule(lr){2-3}\cmidrule(lr){4-5}\cmidrule(lr){6-7}\cmidrule(lr){8-9}
& \textbf{$\Delta$Acc (\%)} & \textbf{Key Sp. (\%)}
& \textbf{$\Delta$Acc (\%)} & \textbf{Key Sp. (\%)}
& \textbf{$\Delta$Acc (\%)} & \textbf{Key Sp. (\%)}
& \textbf{$\Delta$Acc (\%)} & \textbf{Key Sp. (\%)} \\
\midrule
ARCe & $-1.30$ & 39.78 & $-1.01$ & 39.75 & $-1.56$ & 39.67 & $-2.27$ & 40.98 \\
ARCc & $+0.00$ & 39.65 & $+0.00$ & 39.62 & $-1.11$ & 39.53 & $-0.85$ & 40.84 \\
BQ   & $-0.12$ & 40.20 & $-0.12$ & 40.18 & $-0.83$ & 40.20 & $-0.12$ & 42.46 \\
HS   & $-0.41$ & 40.86 & $-0.63$ & 40.83 & $-0.80$ & 40.83 & $-0.77$ & 42.37 \\
OQ   & $+1.00$ & 38.90 & $-0.20$ & 38.88 & $-0.80$ & 38.91 & $-1.20$ & 41.02 \\
PQ   & $-1.25$ & 39.79 & $-1.09$ & 39.78 & $-1.52$ & 39.67 & $-2.01$ & 41.26 \\
WG   & $-1.10$ & 39.58 & $-0.55$ & 39.55 & $-0.79$ & 39.58 & $-0.95$ & 42.15 \\
\midrule
\textbf{Average} & \textbf{$-0.46$} & \textbf{39.82} & \textbf{$-0.52$} & \textbf{39.80} & \textbf{$-1.06$} & \textbf{39.77} & \textbf{$-1.17$} & \textbf{41.58} \\
\bottomrule
\end{tabular*}
\endgroup
\end{table*}

Table~\ref{tab:weight_compression} shows that the average accuracy changes are $-0.46$, $-0.52$, $-1.06$, and $-1.17$ percentage points for BF16, INT8 RTN, 4-bit AWQ, and 50\% Wanda pruning, respectively. The corresponding average Delta-Key sparsities are 39.82\%, 39.80\%, 39.77\%, and 41.58\%.
INT8 RTN provides an accuracy--sparsity trade-off close to that of the BF16 model, while the more aggressive AWQ and Wanda configurations incur slightly larger but still moderate accuracy degradation. Overall, DeltaLLM remains effective across the different weight representations with comparable sparsity.

\subsection{Speedup and Resource Utilization Analysis}

We first distinguish the speedup contributed by the sparse-attention algorithms from that contributed by the specialized DeltaLLM dataflow and hardware using the resource-matched CVE described in Section~\ref{sec:setup}.

\begin{table}[t]
\centering
\caption{
Resource-matched speedup over the dense attention baseline on the CVE simulator across input lengths. 
The first four configurations use the generic CVE mapping, whereas DeltaLLM uses the specialized architecture. }
\label{tab:cve_comparison}
\begingroup

\footnotesize
\setlength{\tabcolsep}{2pt}
\renewcommand{\arraystretch}{1.12}

\begin{tabular*}{\columnwidth}{
@{\extracolsep{\fill}}
>{\raggedright\arraybackslash}m{0.45\columnwidth}
ccccc@{}
}
\toprule
& \multicolumn{5}{c}{\textbf{Input length}} \\
\cmidrule(lr){2-6}
\textbf{Configuration}
& \textbf{128}
& \textbf{256}
& \textbf{512}
& \textbf{1{,}024}
& \textbf{2{,}048} \\
\midrule

StreamingLLM (CVE)
& \textbf{2.40$\times$}
& \textbf{2.44$\times$}
& \textbf{2.46$\times$}
& \textbf{2.49$\times$}
& 2.49$\times$ \\

MInference (CVE)
& 1.09$\times$
& 1.50$\times$
& 1.86$\times$
& 2.13$\times$
& 2.29$\times$ \\

$\Delta K$ + Hybrid (CVE)
& 1.14$\times$
& 1.15$\times$
& 1.17$\times$
& 1.17$\times$
& 1.18$\times$ \\

$\Delta K$ + Hybrid + VSSOS (CVE)
& 1.81$\times$
& 1.89$\times$
& 1.97$\times$
& 2.05$\times$
& 2.12$\times$ \\

\addlinespace[2pt]
\textbf{DeltaLLM (specialized)}
& 2.23$\times$
& 2.33$\times$
& 2.42$\times$
& \textbf{2.49$\times$}
& \textbf{2.55$\times$} \\

\bottomrule
\end{tabular*}
\endgroup
\end{table}

As shown in Table~\ref{tab:cve_comparison}, when mapped onto the 
generic CVE, the combination of Delta-Key and hybrid attention achieves 1.14$\times$--1.18$\times$ speedup over the dense attention baseline simulated on the same CVE. Adding VSSOS increases the speedup to 1.81$\times$--2.12$\times$, demonstrating the importance of reducing the score--value path after the query--key computation has been sparsified. Therefore, the DeltaLLM algorithm itself provides 1.81$\times$-2.12$\times$ speedup. Mapping the complete algorithm onto the specialized DeltaLLM dataflow and SKG/VPU/SOSU architecture further increases the speedup to 2.23$\times$--2.55$\times$. 
This comparison shows that the overall acceleration results from both the proposed sparsity mechanisms and their specialized dataflow and hardware implementation.

For comparison, we also evaluate StreamingLLM and MInference using the same resource-matched CVE setup.
StreamingLLM achieves high simulated speedup at shorter input lengths because its fixed pattern aggressively removes attention entries.
However, as shown in Table~\ref{tab:deltallm_minference_streaming_avg}, this comes with substantial accuracy degradation. MInference preserves accuracy more effectively than StreamingLLM but incurs runtime pattern-generation overhead, particularly at shorter lengths. 
At an input length of 2{,}048, the complete DeltaLLM design achieves 2.55$\times$ speedup, compared with 2.49$\times$ for StreamingLLM and 2.29$\times$ for MInference, while providing a better accuracy--sparsity trade-off.

We next compare the physical FPGA implementation of DeltaLLM with a SOTA prefilling-attention accelerator.
Among reported edge-FPGA designs, TeLLMe-v2~\cite{qiao2026tellme} provides the most directly comparable prefilling-attention latency and resource results.
TeLLMe-v2 runs an end-to-end BitNet-0.7B model on the KV260 at 250\,MHz with a dedicated reverse prefilling attention (RPA) unit to accelerate prefilling attention.
It reports using 6 PEs in RPA, while the actual number of multiplication and exponential units is around 2--3 in each PE.
For a fair comparison, we run the same model on \textit{DeltaLLM} with $N_{pe}=16$, since there is only one multiplier in each of our PEs.
We set $\theta_k=1.2$, $\gamma=0.05$ and $T_s=-10$ for the performance evaluation.
Table~\ref{tab:latency_comparison} shows the prefilling latency when the input length is 128.
DeltaLLM reduces the reported per-layer prefilling-attention latency from 7.6~ms for TeLLMe-v2 to 4.7~ms, corresponding to a 1.62$\times$ speedup.

\begin{table*}[!t]
\caption{FPGA-based comparison for one layer of LLM prefilling attention.}
\label{tab:latency_comparison}
\centering
\footnotesize
\setlength{\tabcolsep}{3pt}
\renewcommand{\arraystretch}{0.9}
\begin{threeparttable}
\resizebox{\textwidth}{!}{%
\begin{tabular}{lccccccccccccc}
\toprule
\multirow{2}{*}{\textbf{Method}} &
\multirow{2}{*}{\textbf{Model}} &
\multirow{2}{*}{\textbf{Platform}} &
\multirow{2}{*}{\textbf{Length}} &
\multirow{2}{*}{\textbf{Precision}} &
\multirow{2}{*}{\textbf{Power (W)}} &
\multicolumn{5}{c}{\textbf{FPGA Resource Utilization}} &
\multicolumn{3}{c}{\textbf{Performance}} \\
\cmidrule(lr){7-11} \cmidrule(lr){12-14}
& & & & & &
\textbf{LUT} & \textbf{FF} & \textbf{DSP} & \textbf{BRAM} & \textbf{URAM} &
\textbf{Latency (ms)} & \textbf{Token/s} & \textbf{Speedup} \\
\midrule

TeLLMe-v2~\cite{qiao2026tellme}
& BitNet-0.7B & KV260 & 128 & FP16 & 4.8\tnote{a}
& 18,072 & 31,212 & 115 & 43.5 & 4
& 7.6 & 16.8k & 1.00$\times$ \\

\textbf{DeltaLLM (Ours)}
& BitNet-0.7B & KV260 & 128 & FP16 & \textbf{2.9}
& \textbf{22,730} & \textbf{32,604} & \textbf{80} & \textbf{4} & \textbf{4}
& \textbf{4.7} & \textbf{27.2k} & \textbf{1.62$\times$} \\
\bottomrule
\end{tabular}%
}
\begin{tablenotes}[flushleft]
\footnotesize
\item[a] The power of TeLLMe-v2 is an end-to-end LLM inference power including the hardware for the linear layers.
\end{tablenotes}
\end{threeparttable}
\end{table*}

\begin{figure}[!t]
\centering
\resizebox{\linewidth}{!}{%
\begin{tikzpicture}

\begin{axis}[
    width=0.95\linewidth,
    height=6.2cm,
    xlabel={Input Length},
    ylabel={Speedup ($\times$)},
    symbolic x coords={128, 256, 512, 1024, 2048},
    xtick=data,
    enlarge x limits=0.08,
    ymin=1.5,
    ymax=2.2,
    ytick={1.5, 1.6, 1.7, 1.8, 1.9, 2.0, 2.1, 2.2},
    axis y line*=left,
    axis x line*=bottom,
    ylabel style={
        font=\small,
        color=blue!70!black
    },
    yticklabel style={
        color=blue!70!black
    },
    label style={
        font=\small
    },
    tick label style={
        font=\small
    },
    ymajorgrids=true,
    grid style={
        dashed,
        gray!25
    },
    legend style={
        at={(0.02,0.98)},
        anchor=north west,
        font=\small,
        draw=none,
        fill=white,
        fill opacity=0.85,
        text opacity=1,
        legend cell align=left
    },
]

\addplot[
    color=blue!70!black,
    mark=*,
    mark size=2.5pt,
    line width=1.2pt,
    point meta=explicit symbolic,
    nodes near coords,
    every node near coord/.append style={
        font=\scriptsize,
        color=blue!70!black,
        anchor=south,
        yshift=4pt,
        fill=white,
        fill opacity=0.85,
        text opacity=1,
        inner xsep=1.5pt,
        inner ysep=1pt,
        rounded corners=1pt
    },
] coordinates {
    (128, 1.75)  [1.75$\times$]
    (256, 1.83)  [1.83$\times$]
    (512, 1.94)  [1.94$\times$]
    (1024, 2.01) [2.01$\times$]
    (2048, 2.13) [2.13$\times$]
};

\addlegendentry{Speedup}

\addlegendimage{
    red!70!black,
    mark=square*,
    line width=1.2pt
}
\addlegendentry{Avg. Sparsity}

\end{axis}

\begin{axis}[
    width=0.95\linewidth,
    height=6.2cm,
    symbolic x coords={128, 256, 512, 1024, 2048},
    xtick=data,
    enlarge x limits=0.08,
    ymin=0.45,
    ymax=0.68,
    ytick={0.45, 0.50, 0.55, 0.60, 0.65},
    axis y line*=right,
    axis x line=none,
    ylabel={Avg. Sparsity},
    ylabel style={
        font=\small,
        color=red!70!black
    },
    yticklabel style={
        color=red!70!black
    },
    tick label style={
        font=\small
    },
]

\addplot[
    color=red!70!black,
    mark=square*,
    mark size=2.5pt,
    line width=1.2pt,
    point meta=explicit symbolic,
    nodes near coords,
    every node near coord/.append style={
        font=\scriptsize,
        color=red!70!black,
        anchor=north,
        yshift=-4pt,
        fill=white,
        fill opacity=0.85,
        text opacity=1,
        inner xsep=1.5pt,
        inner ysep=1pt,
        rounded corners=1pt
    },
] coordinates {
    (128, 0.51)  [0.51]
    (256, 0.54)  [0.54]
    (512, 0.58)  [0.58]
    (1024, 0.60) [0.60]
    (2048, 0.63) [0.63]
};

\end{axis}
\end{tikzpicture}%
}
\caption{
Scalability of DeltaLLM at input lengths from 128 to
2{,}048. Speedup is normalized to dense full attention using a
FlashAttention-style~\cite{dao2022flashattention} dataflow on the same
hardware; average sparsity includes the two attention matrix
multiplications.
}
\label{fig:speedup_sparsity}
\end{figure}

The input length of 128 in Table~\ref{tab:latency_comparison} is dictated only by the comparable latency reported by TeLLMe-v2 and does not limit our scalability evaluation. 
To evaluate scalability on longer input lengths, we implement a dense FlashAttention-style ~\cite{dao2022flashattention} attention on the same DeltaLLM hardware by forcing all blocks to dense mode and bypassing VSSOS, and compare it with DeltaLLM using $\theta_k=1.2$, $\gamma=0.05$, and $T_s=-10$. 
As shown in Fig.~\ref{fig:speedup_sparsity}, the speedup increases from 1.75$\times$ at 128 tokens to 2.13$\times$ at 2{,}048 tokens, while the average sparsity across the two attention matrix multiplications increases from 51\% to 63\%. Thus, the performance benefit is maintained and becomes more pronounced as the input length increases. Table~\ref{tab:cve_comparison} independently confirms this trend for the resource-matched implementation stack over the same 128--2{,}048-token range.

The measured speedup remains below the ideal sparsity-derived limit because some online-softmax vector operations cannot be skipped and the separated vector operations incur memory/register access and pipeline-control costs. These effects are quantified in the sparsity--overhead analysis below.

Table~\ref{tab:resource_utilization} shows the resource-utilization breakdown of \textit{DeltaLLM} and the utilization of TeLLMe-v2 RPA.
Our additional LUT and FF consumption mainly comes from the SKGs and vector registers. 
Moreover, the control unit cost of TeLLMe-v2 RPA is not included.
Importantly, we use many fewer DSP resources than TeLLMe-v2 while achieving lower prefilling latency, owing to the sparsity exploited by our algorithm.
The large BRAM reduction mainly comes from two factors: (1) the dataflow of \textit{DeltaLLM} enables keys and values to stream from DDR without being saved on chip; and (2) unlike TeLLMe-v2, which calculates several attention heads at the same time, \textit{DeltaLLM} computes head by head, so the hardware only stores the memory context related to one attention head.

\begin{table}[!t]
\caption{Resource Utilization Breakdown of DeltaLLM}
\label{tab:resource_utilization}
\centering
\begin{tabular*}{\columnwidth}{@{\extracolsep{\fill}}lrrrrr@{}}
\toprule
\textbf{Module} & \textbf{LUT} & \textbf{FF} & \textbf{DSP} & \textbf{BRAM} & \textbf{URAM} \\
\midrule
SKG                    & 1,968  & 3,008  & 8  & 0 & 0 \\
VPU                    & 18,128 & 21,248 & 72 & 0 & 0 \\
Memory Units           & 850    & 6182   & 0  & 4 & 4\\
Control                & 1,298  & 1,510  & 0  & 0 & 0\\
Misc                   & 486    & 656    & 0  & 0 & 0 \\
\midrule
\textbf{DeltaLLM Total} & \textbf{22,730} & \textbf{32,604} & \textbf{80} & \textbf{4} & \textbf{4} \\
\textbf{TeLLMe-v2 RPA}  &\textbf{18,072} &\textbf{31,212} &\textbf{115} &\textbf{43.5} &\textbf{4} \\
\bottomrule
\end{tabular*}
\end{table}

\subsection{Overhead Analysis}

There are mainly three sources of the latency overhead in DeltaLLM: (1) The Delta-Key path introduces $O(nd_h)$ FP16 additions for Delta-Key generation; (2) $O(n^2)$ additions for recursive score recovery and (3) SKG FIFO-arbitration scheduling overhead. 

Regarding the memory overhead, the sparse buffering cost is small. With $d_h=64$, 16-bit values, and 8-bit indices, the Delta-Key FIFO capacity is $64\times2=128$~B, the reference-key FIFO requires another 128~B, and the address FIFO requires $64\times1=64$~B, for a total of 320~B.


\begin{table}[t]
\centering
\caption{Prefilling-attention cycle breakdown of DeltaLLM at 60\% target Delta-Key sparsity (VSSOS disabled).}
\label{tab:latency_breakdown}
\begingroup
\footnotesize
\setlength{\tabcolsep}{1.5pt}
\renewcommand{\arraystretch}{1.10}
\begin{tabular*}{\columnwidth}{@{\extracolsep{\fill}}lcccccc@{}}
\toprule
& \multicolumn{6}{c}{\textbf{Fraction of Total Attention Cycles (\%)}} \\
\cmidrule(lr){2-7}
\textbf{Input Length}
& \textbf{SKG}
& \makecell{\textbf{FIFO}\\\textbf{Arb.}}
& \makecell{\textbf{Score}\\\textbf{Recovery}}
& \textbf{$QK^{\top}$}
& \makecell{\textbf{Softmax}\\\textbf{+ $SV$}}
& \textbf{Other} \\
\midrule
128     & 1.71\% & 1.14\% & 1.45\% & 29.94\% & 61.77\% & 3.99\% \\
256     & 0.88\% & 0.61\% & 1.47\% & 29.04\% & 64.46\% & 3.54\% \\
512     & 0.53\% & 0.38\% & 1.46\% & 30.23\% & 64.14\% & 3.26\% \\
1{,}024 & 0.31\% & 0.22\% & 1.45\% & 30.41\% & 64.49\% & 3.12\% \\
2{,}048 & 0.17\% & 0.14\% & 1.43\% & 30.33\% & 64.86\% & 3.07\% \\
\bottomrule
\end{tabular*}
\endgroup
\end{table}


Table~\ref{tab:latency_breakdown} further shows the cycle breakdown of DeltaLLM components at 60\% key sparsity while VSSOS is disabled for different input lengths. The first three columns (SKG, FIFO-arbitration, and score-recovery) are the overhead generated by DeltaLLM, while the $QK^{\top}$ and Softmax $+SV$ columns accounts most of the cycles. The 'Other' column mainly consists of the tile setup overhead and the normalization operation. The percentage of SKG, FIFO-arbitration and score-recovery decreases from 4.30\% at 128 tokens to 1.74\% at 2{,}048 tokens. In particular, FIFO arbitration accounts for only 1.14\% at 128 tokens and 0.14\% at 2{,}048 tokens. This decrease is expected because Delta-Key generation and per-token pipeline filling grow linearly with the input length, whereas the dominant attention computation grows quadratically.
Therefore, the measured overhead remains small relative to the main matrix-computation stages, and its relative contribution decreases as the input length increases.

\subsection{Algorithmic Component Ablation Study}
While Table~\ref{tab:cve_comparison} separates the generic and specialized implementation stacks under resource-matched assumptions, this subsection isolates the contribution of each DeltaLLM algorithmic component on our proposed hardware.
\begin{table}[t]
\centering
\caption{DeltaLLM component ablation on BitNet-b1.58-2B-4T using the proposed hardware.}
\label{tab:module_ablation}
\setlength{\tabcolsep}{4pt}
\resizebox{\columnwidth}{!}{%
\begin{threeparttable}
\small
\begin{tabular}{lcccc}
\toprule
\textbf{Config.} & \textbf{Acc.} & \textbf{Key Sp.} & \textbf{Score Sp.} & \textbf{Speedup} \\
\midrule
Baseline\tnote{a}  & 80.48\% & 0\%  & 0\%  & 1.00$\times$ \\
+ $\Delta K$ ($\theta_k = 1.2$)                & 78.56\% & 74\% & 0\%  & 1.45$\times$ \\
+ Hybrid ($\gamma = 0.05$)                 & 79.64\% & 65\% & 0\%  & 1.38$\times$ \\
+ VSSOS (DeltaLLM)             & 79.64\% & 65\% & 63\% & \textbf{2.18$\times$} \\
\midrule
VSSOS only ($T_s=-10$) & 80.48\% & 0\% & 62\% & 1.50$\times$ \\
\bottomrule
\end{tabular}
\begin{tablenotes}
\small
\item[a] Baseline performs FlashAttention-style dense attention without any sparsification on the same hardware.
\end{tablenotes}
\end{threeparttable}
}%
\end{table}

The ablation study is performed on the BoolQ~\cite{clark2019boolq} dataset described in Section~\ref{sec:setup}.
As reported in Table~\ref{tab:module_ablation},
applying Delta-Key generation alone ($\theta_k = 1.2$) achieves 74\% key sparsity and a 1.45$\times$ speedup, but degrades accuracy by 1.92 points because the
approximation is applied indiscriminately across the attention map, including salient regions near the diagonal. 
Introducing the context-aware hybrid strategy with $\gamma=0.05$ recovers 1.08 of these points by preserving full attention within the local context window; the speedup correspondingly drops to 1.38$\times$ since more blocks require exact
computation. 
Finally, adding VSSOS ($T_s = -10$) contributes 63\% structured sparsity in the score-value multiplication at no additional accuracy cost, raising the speedup to 2.18$\times$. 
The independent VSSOS-only configuration in Table~\ref{tab:module_ablation} keeps the query-key computation dense, preserves the baseline BoolQ accuracy of 80.48, produces 62\% score sparsity, and achieves a 1.50$\times$ speedup, close to the 1.45$\times$ speedup of Delta-Key alone.


\section{Conclusion}

We presented \textit{DeltaLLM}, a training-free framework that exploits Delta-Key sparsity to accelerate LLM prefilling attention on edge devices. By combining causal-attention-aware Delta-Key construction, context-aware hybrid attention, and a hardware-friendly sparse online softmax algorithm, \textit{DeltaLLM} effectively exploits temporal and spatial sparsity while maintaining competitive accuracy.
Supported by an optimized hardware architecture and dataflow, \textit{DeltaLLM} achieves substantial performance improvements under limited resource budgets.
On the KV260, it reduces the reported per-layer prefilling-attention latency from 7.6~ms for TeLLMe-v2 to 4.7~ms, corresponding to a 1.62$\times$ speedup while using fewer DSP and BRAM resources.
These results demonstrate the potential of Delta-Key sparse attention as an efficient solution for edge LLM deployment.
Future work will investigate a decode-optimized sparse-attention path and larger-scale LLMs.



\bibliographystyle{IEEEtran}
\bibliography{references}

\begin{IEEEbiography}[{\includegraphics[width=1in,height=1.25in,clip,keepaspectratio]{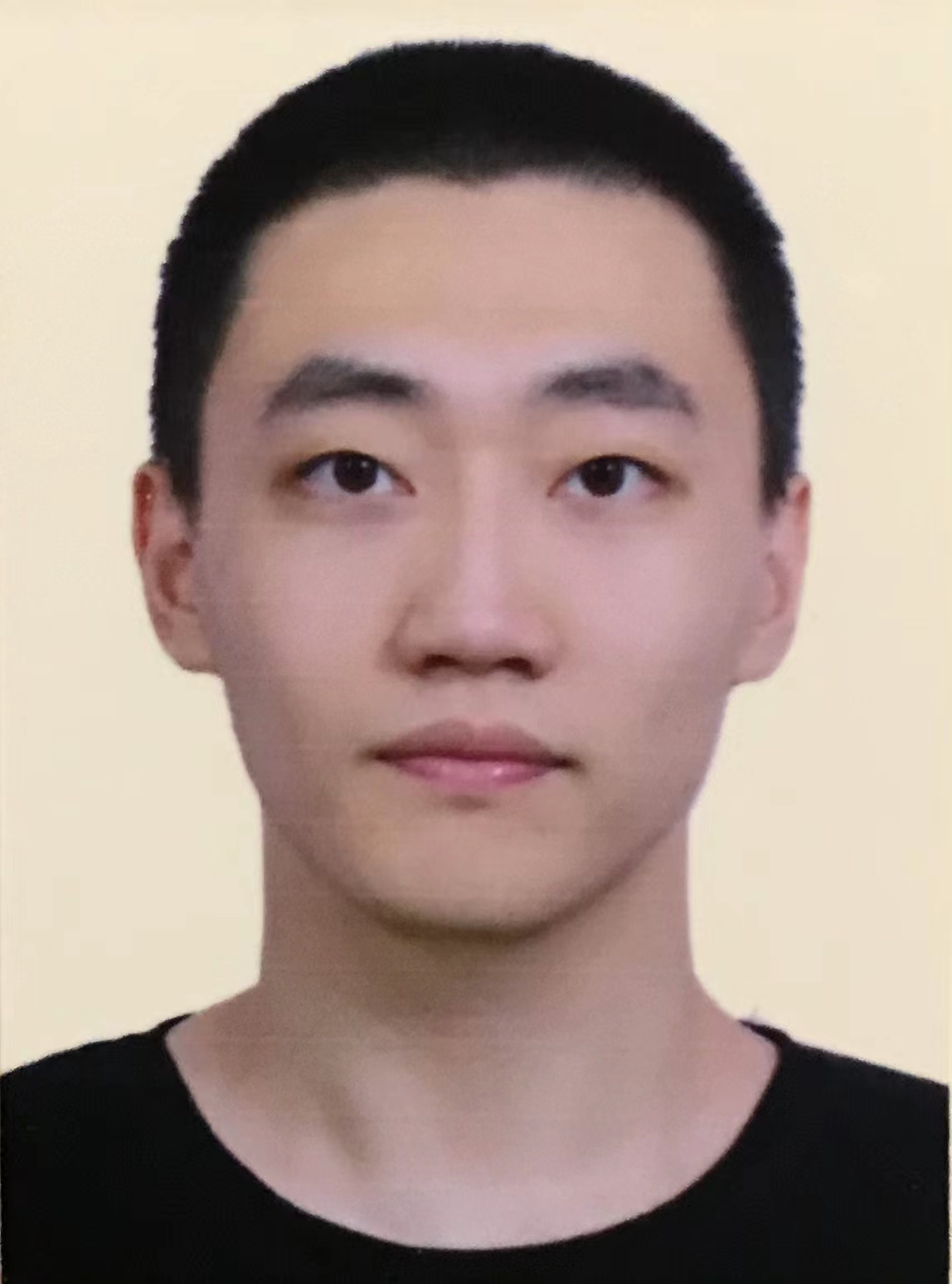}}]{Jiawen Qi}
received his B.Sc.\ degree in Electrical Engineering from the University of New South Wales, Sydney, NSW, Australia, in 2021, and the M.Sc.\ degree in Communication Engineering from the Technical University of Munich, Munich, Germany, in 2024. He has been working toward the Ph.D.\ degree with the Leiden Institute of Advanced Computer Science (LIACS), Leiden University, Leiden, The Netherlands, since 2024.
His current research interests include efficient large language model (LLM) inference and algorithm-hardware co-optimization for AI accelerators.
\end{IEEEbiography}

\begin{IEEEbiography}[{\includegraphics[width=1in,height=1.25in,clip,keepaspectratio]{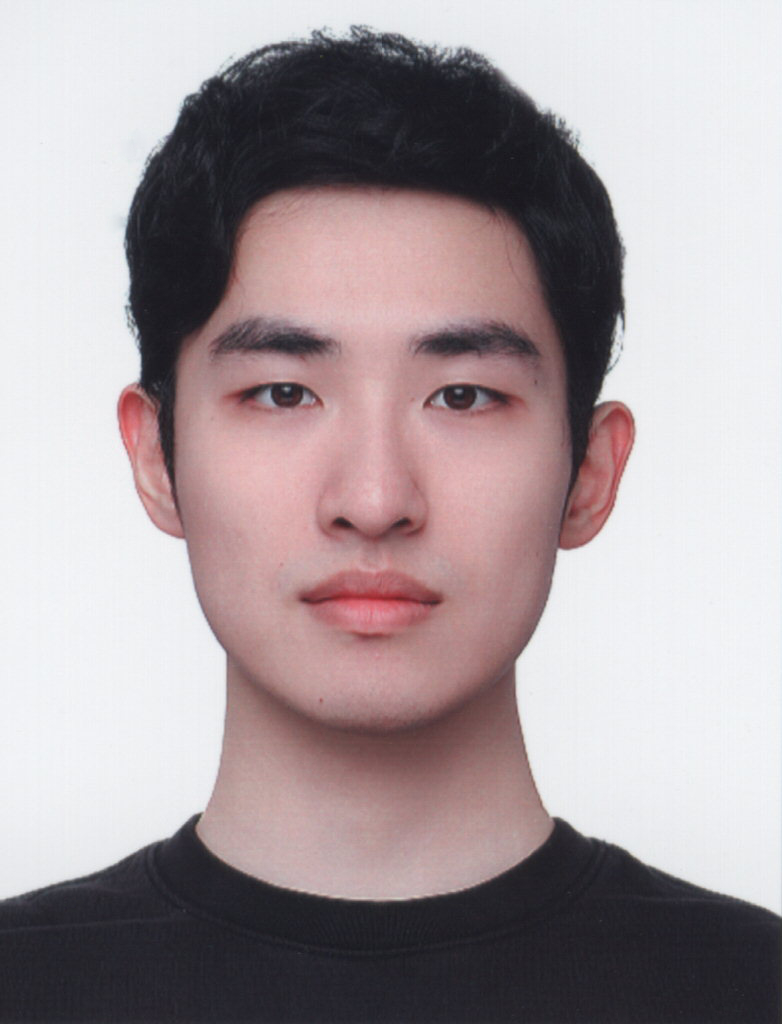}}]{Jiayu Liu}
received his M.Eng. degree in Electronic and Electrical Engineering from University College London in 2025. He is currently pursuing the Ph.D. degree with LIACS, Leiden University, Leiden, The Netherlands. His research interests include audio LLMs, speech therapy systems, and dysarthric speech analysis.
\end{IEEEbiography}

\begin{IEEEbiography}[{\includegraphics[width=1in,height=1.25in,clip,keepaspectratio]{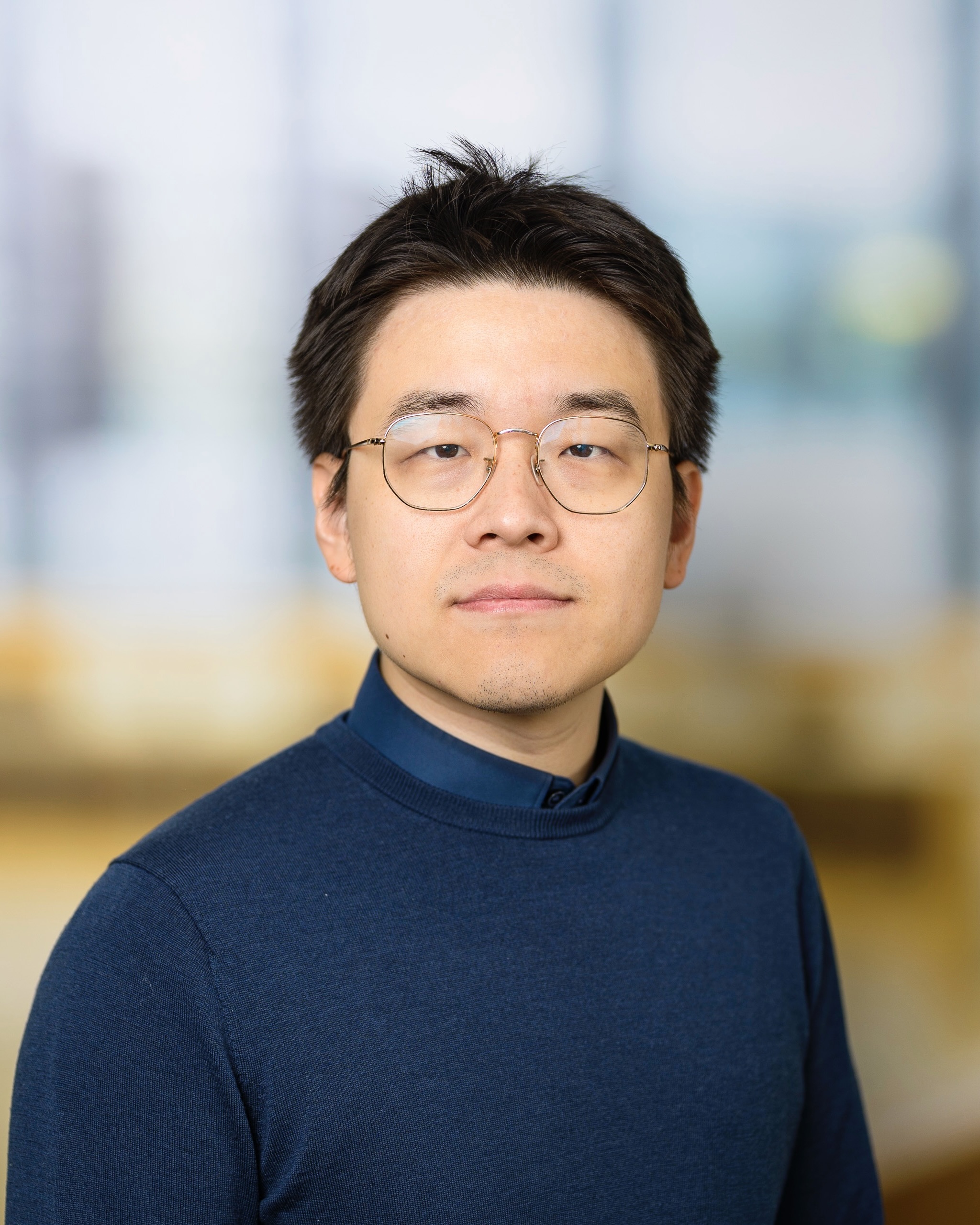}}]{Chang Gao} (Member, IEEE) received his Ph.D. degree with distinction in Neuroscience from the Institute of Neuroinformatics, University of Zürich and ETH Zürich in March 2022, and his M.Sc. degree from Imperial College London in September 2016 and his B.Eng. degree from the University of Liverpool and Xi’an Jiaotong-Liverpool University in July 2015. Since August 2022, he has been an Assistant Professor at Delft University of Technology, The Netherlands, in the Department of Microelectronics, where he leads the Lab of Efficient Machine Intelligence (EMI) and conducts research on algorithm-hardware co-design for edge intelligence.
\end{IEEEbiography}

\begin{IEEEbiography}[{\includegraphics[width=1in,height=1.25in,clip,keepaspectratio]{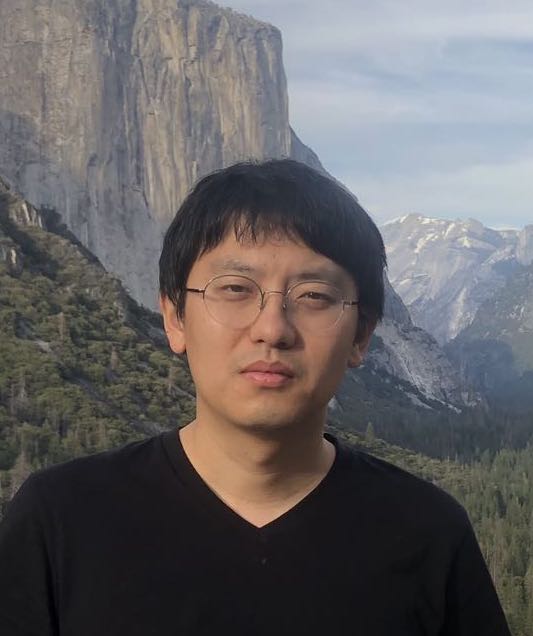}}]{Zhaochun Ren} received the Ph.D. degree in Computer Science from University of Amsterdam, Amsterdam, The Netherlands, in 2016. Zhaochun Ren is an Associate Professor at Leiden University, the Netherlands. He is interested in information retrieval and natural language processing, with an emphasis on generative information retrieval and recommender systems. He aims to develop intelligent agents that can address complex user requests and solve core challenges in NLP and IR towards that goal. His research has been recognized with multiple awards at RecSys, SIGIR, WSDM, EMNLP, and CIKM. Prior to joining Leiden, he was a Professor at Shandong University and a Research Scientist at JD.com.
\end{IEEEbiography}

\begin{IEEEbiography}[{\includegraphics[width=1in,height=1.25in,clip,keepaspectratio]{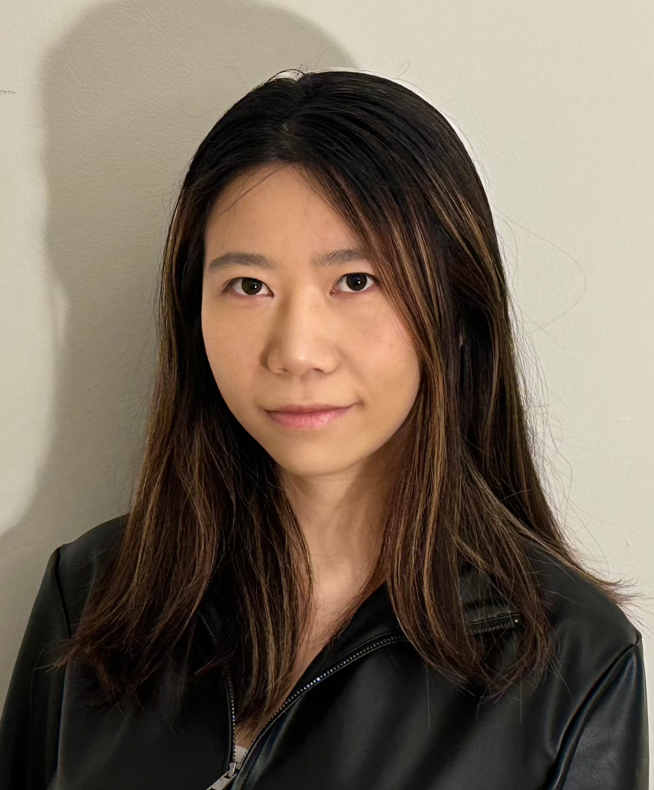}}]{Qinyu Chen} (Senior Member, IEEE) received the Ph.D. degree in Electronic Science and Technology from Nanjing University, Nanjing, China, in 2021, and the B.Eng. degree in Communication Engineering from Shandong University, Jinan, China, in 2016. In 2022, she joined the Institute of Neuroinformatics, University of Zürich and ETH Zürich, Zurich, Switzerland as a postdoc. 
Since 2024, she has been an Assistant Professor at the Leiden Institute of Advanced Computer Science (LIACS), Leiden University, Leiden, the Netherlands. 
Her current research interests include efficient brain-inspired AI systems at the edge. She received a 2022 Bridge Fellowship Grant from the Swiss National Science Foundation (SNSF) and Innosuisse, and a 2024 VENI Talent program from the Dutch Research Council (NWO). She serves as the chair-elect of the Neural Systems and Applications (NSA) Technical Committee in the IEEE Circuits and Systems Society (CASS) and chair of the CAS Benelux chapter. 
\end{IEEEbiography}


 




\vfill

\end{document}